%% file: main.tex
\documentclass[10pt,twocolumn,letterpaper]{article}

\usepackage[pagenumbers]{cvpr} %

\usepackage{graphicx}
\usepackage{amsmath}
\usepackage{amssymb}
\usepackage{booktabs}

\usepackage[pagebackref,breaklinks,colorlinks]{hyperref}
\newcommand\customparagraph[1]{\vspace{0.4em}\noindent\textbf{#1.}}
\newcommand{\minus}{\,\scalebox{0.8}[1.0]{$-$}\,}
\usepackage{comment}
\usepackage{widetable}
\usepackage{multirow}
\usepackage{color}

\usepackage[capitalize]{cleveref}
\usepackage{amsmath}
\DeclareMathOperator*{\argmax}{arg\,max}

\crefname{section}{Sec.}{Secs.}
\Crefname{section}{Section}{Sections}
\Crefname{table}{Table}{Tables}
\crefname{table}{Tab.}{Tabs.}
\crefname{equation}{Eq.}{Eqs.}
\newcommand{\Eq}{Eq.\@\xspace}

\begin{document}

\input{tex/title}
\input{tex/abstract}
\input{tex/introduction}
\input{tex/related_work}

\input{tex/method}

\input{tex/experiments}
\input{tex/conclusion}

\input{tex/appendix}

\clearpage

{\small
\bibliographystyle{ieee_fullname}
\bibliography{main}
}

\end{document}

%% file: tex/title.tex
\title{IterMVS: Iterative Probability Estimation for Efficient Multi-View Stereo}

\author{Fangjinhua Wang$^1$
        \quad
        Silvano Galliani$^2$
        \quad
        Christoph Vogel$^2$
        \quad
        Marc Pollefeys$^{1,2}$\\
        $^1$Department of Computer Science, ETH Zurich\\
        $^2$Microsoft Mixed Reality \& AI Zurich Lab}
        
\maketitle

%% file: tex/abstract.tex
\begin{abstract}
We present IterMVS, a new data-driven method for high-resolution multi-view stereo. 
We propose a novel GRU-based estimator that encodes pixel-wise probability distributions of depth in its hidden state. 
Ingesting multi-scale matching information, our model refines these distributions over multiple iterations and infers depth and confidence. 
To extract the depth maps, %
we combine traditional classification and regression in a novel manner. 
We verify the efficiency and effectiveness of our method on DTU, Tanks\&Temples and ETH3D.
While being the most efficient method in both  memory and run-time, our model achieves competitive performance on DTU and better generalization ability on Tanks\&Temples as well as ETH3D than most state-of-the-art methods. 
Code is available at \url{https://github.com/FangjinhuaWang/IterMVS}.

\end{abstract}

%% file: tex/introduction.tex
\section{Introduction}
\label{sec:intro}

Multi-view stereo (MVS) describes the technology to reconstruct dense 3D models of observed scenes from a set of calibrated images. 
MVS is a fundamental problem of geometric computer vision and a core technique for applications like augmented/virtual reality, autonomous driving and robotics.
Albeit being studied extensively for decades, the conditions that occur in real-world application scenarios pose problems such as occlusion, illumination changes, low-textured areas and non-Lambertian 
surfaces~\cite{aanaes2016_dtu,knapitsch2017tanks,2017eth3d,yao2020blendedmvs} that remain unsolved up to now.

\begin{figure}[tbp]
\vspace{-0.2cm}
\setlength{\belowcaptionskip}{-0.6cm}
\centering
\setlength{\abovecaptionskip}{-0.2cm}
{\includegraphics[width=1.0\linewidth]{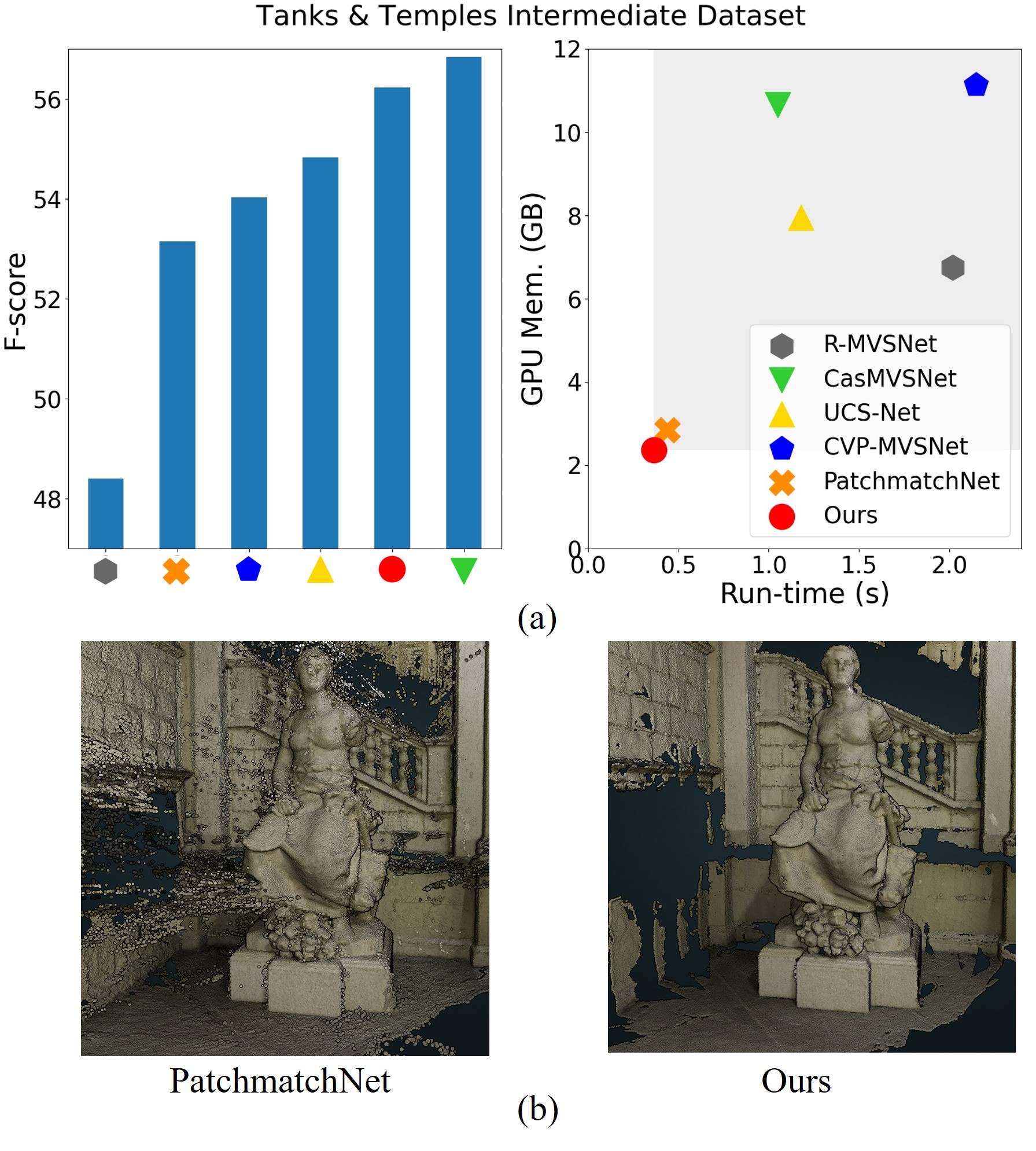}}
\caption{
(a) Comparison with state-of-the-art learning-based MVS 
methods~\cite{yao_2019_rmvsnet,gu_2020_cascademvsnet,cheng_2020_ucsnet,yang_2020_cvpr, patchmatchnet_wang} 
on Tanks \& Temples~\cite{knapitsch2017tanks} (all methods are trained on DTU~\cite{aanaes2016_dtu} only).
\textbf{Left}: F-score ($\uparrow$). \textbf{Right}: GPU memory and run-time (image size $1920\!\times\!1024$, 7 views).
(b) Qualitative comparison with PatchmatchNet on ETH3D~\cite{2017eth3d}. Our reconstruction contains significantly less noise. 
}\label{fig:teaser}
\end{figure}

Traditional methods~\cite{galliani_2015_gipuma, schoenberger2016colmap, xu_2019_acmm, furukawa2010} suffer from hand-crafted modeling and matching metrics and struggle under those challenging conditions. 
In comparison, recent data-driven approaches~\cite{yao_2018_mvsnet,yao_2019_rmvsnet,yu_2020_fastmvsnet,gu_2020_cascademvsnet,patchmatchnet_wang} based on Convolutional Neural Networks (CNN) 
demonstrate a significantly improved performance on various MVS benchmarks~\cite{aanaes2016_dtu,knapitsch2017tanks}. 

A popular representative, MVSNet~\cite{yao_2018_mvsnet}, constructs a 3D cost volume that is regularized with a 3D CNN and regresses the depth map from the probability volume. 
While this methodology achieves impressive performance on benchmarks, it does not scale up well to high-resolution images or large-scale scenes, because the 3D CNN is memory and run-time consuming. %
However, low run-time and power consumption are the key in most industrial applications 
and resource friendly methodologies become more important. %
Aiming to improve the efficiency, recent variants of MVSNet~\cite{yao_2018_mvsnet} are proposed, which can be mainly divided into two categories: 
recurrent methods~\cite{yao_2019_rmvsnet, d2hcrmvsnet} and multi-stage methods~\cite{gu_2020_cascademvsnet, cheng_2020_ucsnet, yang_2020_cvpr, patchmatchnet_wang}. 
Several recurrent methods~\cite{yao_2019_rmvsnet, d2hcrmvsnet} can relax the memory consumption by regularizing the cost volume sequentially with GRU~\cite{cho2014learning} or convolutional LSTM~\cite{xingjian2015convolutional}, however, at the cost of an increase in run-time. 
In contrast, multi-stage methods~\cite{gu_2020_cascademvsnet, cheng_2020_ucsnet, yang_2020_cvpr} utilize cascade cost volumes and estimate the depth map from coarse to fine. 
While this approach can lead to high efficiency in both memory and run-time, a reduced search range at finer stages implies a limitation in recovering from errors induced at coarse resolutions~\cite{teed2020raft}.

A current method that unites a competitive performance with highest efficiency in memory and run-time among all learning-based methods is PatchmatchNet~\cite{patchmatchnet_wang}. 
Based on traditional PatchMatch~\cite{galliani_2015_gipuma, bleyer2011patchmatch}, PatchmatchNet combines learned adaptive propagation and evaluation modules with a cascaded structure. 
While sharing the common limitation of coarse-to-fine methods, the generalization ability of PatchmatchNet appears limited when compared to other multi-stage methods~\cite{gu_2020_cascademvsnet, cheng_2020_ucsnet, yang_2020_cvpr}.

In this work, we propose IterMVS, a novel GRU-based iterative method aimed at further improving efficiency as well as performance for high-resolution MVS. 

\vspace{0.4em}\noindent\textbf{Contributions:}
\textbf{(\romannumeral 1)}
We propose a novel and lightweight GRU-based probability estimator that encodes the per-pixel probability distribution of depth in its hidden state. 
This compressed representation does not require to keep the probability volume in memory the whole time. 
In each iteration, multi-scale matching information is injected to update the pixel-wise depth distribution. 
Compared to coarse-to-fine methods, the GRU-based probability estimator always operates at the same resolution, utilizes a large search range and keeps track of the distribution over the full depth range.
\textbf{(\romannumeral 2)}
We propose a simple, yet effective depth estimation strategy that combines both classification and regression, which is robust to multi-modal distributions but also achieves sub-pixel precision.
\textbf{(\romannumeral 3)}
We verify the effectiveness of our method on various MVS datasets, \eg DTU~\cite{aanaes2016_dtu}, Tanks \& Temples~\cite{knapitsch2017tanks} and ETH3D~\cite{2017eth3d}. 
The results demonstrate that IterMVS achieves very competitive performance, while showing highest efficiency in both memory and run-time among all the learning-based methods, Fig.~\ref{fig:teaser}. 
Compared with PatchmatchNet~\cite{patchmatchnet_wang}, IterMVS is more efficient in both memory and run-time, achieves comparable performance on DTU~\cite{aanaes2016_dtu} and demonstrates much better generalization ability on Tanks \& Temples~\cite{knapitsch2017tanks} and ETH3D~\cite{2017eth3d}.

%% file: tex/related_work.tex
\section{Related Work}
\customparagraph{Traditional MVS}
Based on the scene representations, traditional MVS methods can be divided into three main categories: 
volumetric, point cloud based and depth map based. 
Volumetric methods~\cite{ulusoy2017, kutulakos2000theory, seitz1999photorealistic, kostrikov2014probabilistic} discretize 3D space into voxels and label each as inside or outside of the true surface. 
Operating in scene space usually comes at the price of large memory and run-time consumption, limiting applications to scenes of smaller scale. 
Point cloud based methods~\cite{lhuillier2005quasi, furukawa2010} operate directly on 3D points and often employ propagation to gradually densify the reconstruction. 
By decoupling the problem into depth map estimation and fusion, depth map based methods~\cite{galliani_2015_gipuma, schoenberger2016colmap, xu_2019_acmm,xu2020acmp} are more concise and flexible. 
Galliani~\etal~\cite{galliani_2015_gipuma} propose Gipuma, a multi-view extension of Patchmatch stereo, which uses a red-black checkerboard pattern to parallelize propagation.   
In COLMAP, Sch\"{o}nberger~\etal~\cite{schoenberger2016colmap} jointly estimate pixel-wise view selection, depth map and surface normal. 
Although traditional depth map based methods can achieve impressive results, hand-crafted models and features limit the performance under challenging conditions.

\customparagraph{Data-driven MVS}
Recently, data-driven methods dominate the research for MVS. 
Several volumetric methods~\cite{ji_2017_surfacenet,kar2017learning} first compute a cost volume from multiple images and infer surface voxels after cost volume regularization  with a 3D CNN. However, similar to traditional volumetric methods, they are restricted to small-scale reconstructions. 
More common are depth map based methods~\cite{yao_2018_mvsnet,chen_2019_pointmvsnet,luo_2019_p_mvsnet,xu2020learning_inverse} that often operate in similar fashion. 
MVSNet~\cite{yao_2018_mvsnet} can be seen as a blueprint. It computes an initial cost volume from the features that is regularized with a 3D CNN and regresses the depth map from the probability volume. 
The high memory consumption of 3D CNNs often limits these methods to down-sampled cost volumes and depth maps.
Recently, several variants based on MVSNet~\cite{yao_2018_mvsnet} were proposed that aim to reduce memory and run-time consumption. 
The two main ideas involve recurrent~\cite{yao_2019_rmvsnet, d2hcrmvsnet} and multi-stage methods~\cite{gu_2020_cascademvsnet, cheng_2020_ucsnet, yang_2020_cvpr, xu2020pvsnetpv, vismvsnet, patchmatchnet_wang}. 
R-MVSNet~\cite{yao_2019_rmvsnet} sequentially regularizes 2D slices of the cost volume with a GRU~\cite{cho2014learning}. 
$D^2$HC-RMVSNet~\cite{d2hcrmvsnet} augments R-MVSNet with a complex convolutional LSTM~\cite{xingjian2015convolutional}. 
The main drawback of these recurrent methods is the high run-time. 
In contrast, multi-stage methods~\cite{gu_2020_cascademvsnet, cheng_2020_ucsnet, yang_2020_cvpr, xu2020pvsnetpv, vismvsnet} 
achieve efficiency in both memory and run-time. 
They operate on cascade cost volumes and estimate the depth map in a coarse-to-fine manner. 
First, a low resolution depth map is computed utilizing a large but coarse sampling interval. 
After upsampling, the estimation is refined at a higher sampling rate but at a smaller interval and search range. 
PatchmatchNet~\cite{patchmatchnet_wang} further proposes an adaptive procedure based on 
PatchMatch~\cite{barnes_2009_patchmatch, bleyer2011patchmatch} that achieves superior efficiency among all learning-based methods. 
Despite their impressive performance, coarse-to-fine methods have difficulty to recover
from errors introduced at coarse resolutions~\cite{teed2020raft}, where the sampling interval is large but the sampling frequency low. 
In contrast, we estimate the depth map at a relatively high resolution and generate hypotheses in a fixed large search range in each GRU iteration.
Further, we let the hidden state of our GRU encode the probability distribution for the whole depth range.

\customparagraph{Iterative Update}
Recently, RAFT~\cite{teed2020raft} proposes to estimate optical flow by iteratively updating a motion field through a GRU, which emulates first-order optimization. 
The idea is further adopted in stereo~\cite{lipson2021raft}, scene flow~\cite{teed2021raft} and SfM~\cite{gu2021dro}.
In our work, we let the GRU model a probability distribution per pixel from which we predict the depth map. 
The hidden state is updated in each iteration to more accurately model the pixel-wise probability distribution. %

%% file: tex/method.tex
\section{Method}
\label{sec:method}
\begin{figure*}[htbp]

\setlength{\belowcaptionskip}{-0.5cm}
\centering
\setlength{\abovecaptionskip}{0.0cm}
{\includegraphics[width=0.778\linewidth]{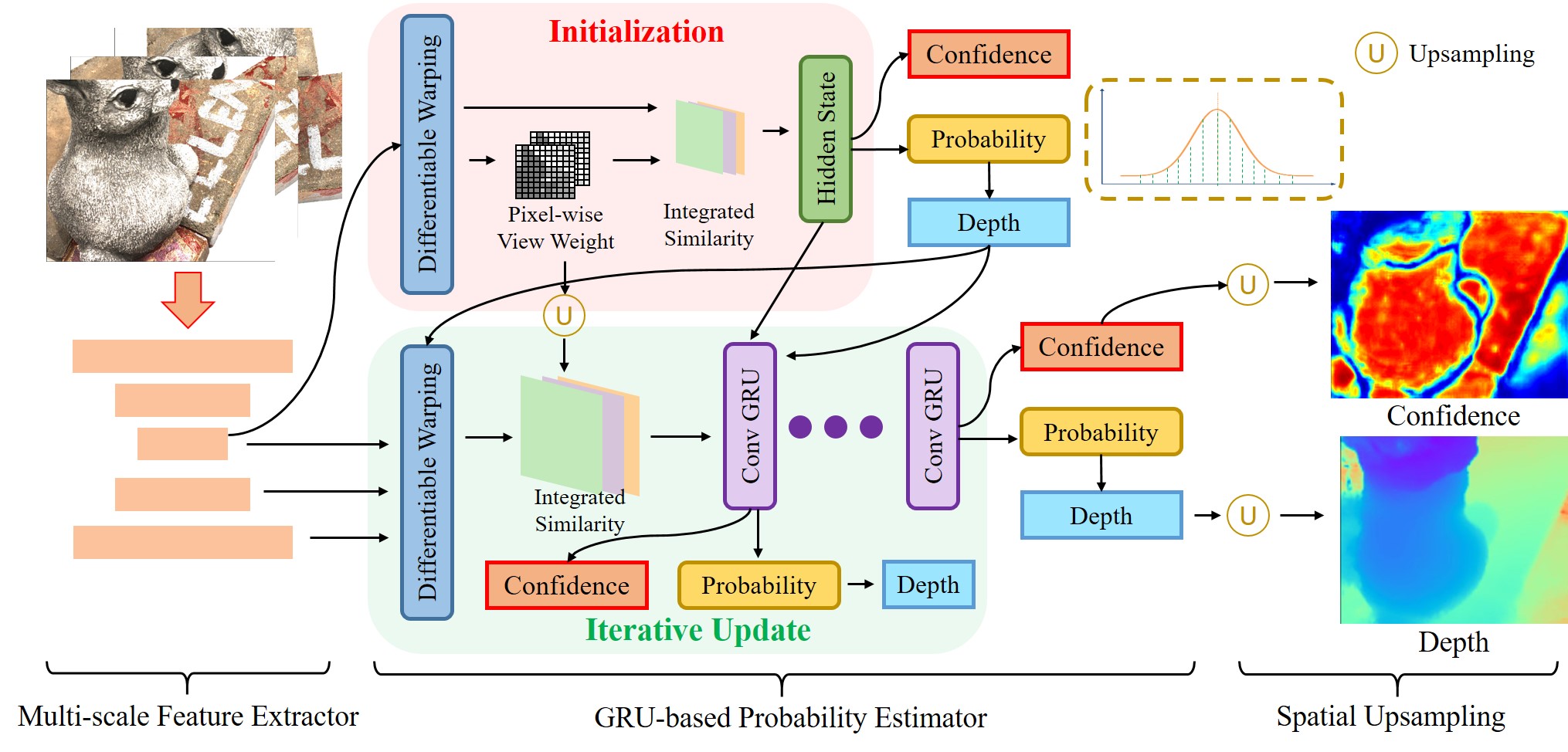}}
\caption{
Detailed structure of IterMVS. 
With multi-scale features extracted from the feature extractor, the GRU-based probability estimator computes matching similarity for depth hypotheses on different scale levels. The GRU takes the matching information as input to iteratively update the hidden state, which encodes the per-pixel probability distributions of depth. %
The depth map and confidence are estimated from the hidden state. %
The spatial upsampling module takes the quarter-resolution depth and confidence output by the GRU to full resolution.
}
\label{fig:pipeline}
\end{figure*}

In this section, we introduce the detailed structure of IterMVS, illustrated in Fig.~\ref{fig:pipeline}.
It consists of a multi-scale feature extractor, an iterative GRU-based probability estimator that models a probability distribution of the depth at each pixel, and a spatial upsampling module. 

\subsection{Multi-scale Feature Extractor}
Given $N$ input images of size $W \!\times\! H$, we use $\mathbf{I}_0$ and ${\left\{\mathbf{I}_i\right\}}_{i=1}^{N-1}$ to denote reference and source images respectively. 
Similar to~\cite{gu_2020_cascademvsnet, patchmatchnet_wang}, we extract multi-scale features from the images with a Feature Pyramid Network (FPN)~\cite{lin2017fpn}. 
We attain features at 3 scale levels, and denote the feature of the $i$-th image at level $l$ by $\mathbf{F}_{i,l}$. 
The features $\mathbf{F}_{i,l}$ are stored at $1/2^l$ resolution and possess $C\!=\!16,32,64$ channels at level $l\!=\!1,2,3$ respectively. 
Although we refrain from using an explicit coarse-to-fine structure, \eg~\cite{gu_2020_cascademvsnet, patchmatchnet_wang}, 
we can include multi-scale context information by using our multi-scale features for matching similarity computation in each iteration of the GRU.
This improves performance as shown in Table~\ref{tab:ablations}. 

\subsection{GRU-based Probability Estimator}
Our core module, the GRU-based probability estimator, models the per-pixel probability distribution of depth with a hidden state of 32 dimensions. 
The GRU operates at $1/4$ resolution, outputs a depth map $\mathbf{D} \in \mathbb{R}^{W/4 \times H/4}$ and is unrolled for $K$ iterations. 

\customparagraph{Differentiable Warping}
Following most learning-based MVS methods~\cite{gu_2020_cascademvsnet, yao_2018_mvsnet, yao_2019_rmvsnet,patchmatchnet_wang}, we warp the source features into front-to-parallel planes \wrt the reference view at the given depth hypotheses. 
Specifically, for a pixel $\mathbf{p}$ in the reference view and the $j$-th depth hypothesis $d_j:=d_j(\mathbf{p})$, with known intrinsic $\left\{\mathbf{K}_i\right\}_{i=0}^{N-1}$ and relative transformations $\left\{\left[\mathbf{R}_{0,i}|\mathbf{t}_{0,i} \right] \right\}_{i=1}^{N-1}$ 
between reference view $0$ and source view $i$, 
we can compute the corresponding pixel $\mathbf{p}_{i,j}:=\mathbf{p}_{i}(d_j)$ in the source view as: 
\begin{equation}
\setlength{\abovedisplayskip}{3pt}
\setlength{\belowdisplayskip}{6pt}
    \mathbf{p}_{i,j} = \mathbf{K}_i \cdot (\mathbf{R}_{0,i}\cdot(\mathbf{K}_0^{-1} \cdot \mathbf{p} \cdot d_j) + \mathbf{t}_{0,i}).
\end{equation}
After de-homogenization, we obtain the warped feature $\mathbf{F}_i(\mathbf{p}_{i,j})$ via bilinear interpolation. 

\customparagraph{2-View Matching Similarity Computation}
Because the principle is same for all feature levels, we omit the subindices denoting the level. 
Given reference and warped $i$-th source feature, $\mathbf{F}_0(\mathbf{p}),\,\mathbf{F}_i(\mathbf{p}_{i,j})\!\in\!\mathbb{R}^C$, we first use group-wise correlation~\cite{guo2019group,xu2020learning_inverse, patchmatchnet_wang} to reduce the dimension. 
By dividing the feature channels evenly into $G\!=\!8$ groups, the $g$-th group similarity 
$\mathbf{s}_{i}(\mathbf{p},j)^g\in\mathbb{R}$ can be computed as:
\begin{equation}\label{eq:similarity}
\setlength{\abovedisplayskip}{3pt}
\setlength{\belowdisplayskip}{3pt}
    \mathbf{s}_{i}(\mathbf{p}, j)^g = \frac{1}{C/G} \left \langle \mathbf{F}_0(\mathbf{p})^g, \mathbf{F}_i(\mathbf{p}_{i,j})^g\right \rangle,
\end{equation}
where $\left \langle \cdot, \cdot \right \rangle$ denotes the dot product.
This results in the similarity $\mathbf{s}_i \in \mathbb{R}^{W \times H \times D \times G}$, where $D$ is the number of depth hypotheses per pixel.

\customparagraph{Initialization}
To initialize the hidden state $h \in \mathbb{R}^{W/4 \times H/4 \times 32}$ of the GRU, we only utilize the features on level 3 and then upsample the results to further reduce the computation. 
Per pixel at $1/8$ resolution and within a pre-defined depth range $[d_{min}, d_{max}]$, we place $D_1$ equidistant depth hypotheses in the \emph{inverse} depth range. Sampling hypotheses uniformly in image space is more suitable for large-scale scenes~\cite{xu2020learning_inverse,yao_2019_rmvsnet}. 
After differentiable warping, we can compute $(N-1)$ 2-view matching similarities $\mathbf{s}_i$ ($i=1, \cdots, N-1$) following \Eq~\ref{eq:similarity}.

For each source view, we further estimate a pixel-wise view weight~\cite{xu2020pvsnetpv, patchmatchnet_wang} 
that provides visibility information and increased robustness when integrating the information from all the source views. 
A lightweight 2D CNN is applied on the image space of $\mathbf{s}_i$ to aggregate local information and reduce the feature channels from $G$ to 1. Applying a \textit{softmax} non-linearity along the depth dimension produces $\mathbf{P}_i \in \mathbb{R}^{W/8 \times H/8 \times D_1}$. For pixel $\mathbf{p}$, the view weight of source view $i$ can then be computed as:
\begin{equation}
\setlength{\abovedisplayskip}{3pt}
\setlength{\belowdisplayskip}{3pt}
    \mathbf{w}_i(\mathbf{p}) = \max \left\{ \mathbf{P}_{i}(\mathbf{p},j) | j=0,1,\dots,D_1-1 \right\}.
\end{equation}
Finally, the integrated matching similarity $\mathbf{S}_\textrm{initial}(\mathbf{p}, j)$ for pixel $\mathbf{p}$ and depth hypothesis $d_j$ is given by:
\begin{equation}
\setlength{\abovedisplayskip}{3pt}
\setlength{\belowdisplayskip}{3pt}
    \mathbf{S}_\textrm{initial}(\mathbf{p},j) = \frac{\sum_{i=1}^{N-1}\mathbf{w}_i(\mathbf{p}) \cdot \mathbf{s}_{i}(\mathbf{p},j)}{\sum_{i=1}^{N-1}\mathbf{w}_i(\mathbf{p})}.
\end{equation}
To also consider spatial correlation of depth maps, we aggregate similarity information from neighboring pixels
by applying a 2D U-Net~\cite{ronneberger2015u} on $\mathbf{S}_\textrm{initial} \in \mathbb{R}^{W/8 \times H/8 \times D_1 \times G}$.
This makes the matching more robust~\cite{yao_2018_mvsnet,gu_2020_cascademvsnet,patchmatchnet_wang}. 
The final convolution layer outputs a 1-channel similarity $\mathbf{\bar{S}}_\textrm{initial} \in \mathbb{R}^{W/8 \times H/8 \times D_1}$. 
Then a 2D CNN, $2\times$ bilinear upsampling and the $\tanh$ nonlinearity are applied sequentially on $\mathbf{\bar{S}}_\textrm{initial}$ to produce the initial hidden state $h_0$.

\customparagraph{Iterative Update}
For each pixel $\mathbf{p}$ at iteration $k$ and level $l$, we generate $N_l$ new depth hypotheses sampled uniformly within an interval of size $2R_l$
in the \textit{normalized inverse} range $[(\frac{1}{\mathbf{D}_{k-1}(\mathbf{p})}-\frac{1}{d_{max}})/(\frac{1}{d_{min}}-\frac{1}{d_{max}})-R_l, (\frac{1}{\mathbf{D}_{k-1}(\mathbf{p})}-\frac{1}{d_{max}})/(\frac{1}{d_{min}}-\frac{1}{d_{max}})+R_l]$, 
centered at the previous depth estimate $\mathbf{D}_{k-1} \in \mathbb{R}^{W/4 \times H/4}$. 
Similar to~\cite{lipson2021raft}, we further ensure that $R_{l-1} < R_l$ to capitalize on the more high-frequent information of the higher resolution features.

We then compute matching similarities $\mathbf{\bar{S}}_k$ from features at all 3 levels to include multi-scale information (see supplementary). 
We use the $2\times$ upsampled pixel-wise view weights to integrate matching similarities from source views and then pass them through a level-wise 2D U-Net to aggregate neighborhood information and reduce the number of channels to 1. 
The final matching similarities for 3 levels are concatenated into $\mathbf{\bar{S}}_k \in \mathbb{R}^{W/4 \times H/4 \times (N_1+N_2+N_3)}$.
Concatenating $\mathbf{D}_{k-1}$ and $\mathbf{\bar{S}}_k$ to $x_k$, we use our convolutional GRU to update the hidden state, 
which further allows the network to propagate information from spatial neighbors:
\vspace{-0.5em}
\begin{equation}
\begin{aligned}
z_k = & \;\sigma(\text{Conv}([h_{k-1}, x_k], W_z)), \\
r_k = & \;\sigma(\text{Conv}([h_{k-1}, x_k], W_r)), \\
\Tilde{h}_k = & \,\tanh(\text{Conv}([r_k \odot h_{k-1}, x_k], W_h)), \\
h_k = & \;(1-z_k) \odot h_{k-1} + z_k \odot \Tilde{h}_k,
\end{aligned}
\end{equation}
where $\sigma(\cdot)$ denotes the \textit{sigmoid} nonlinearity, $\odot$ denotes the Hadamard Product.
With the multi-scale matching information injected in each iteration, the hidden state can encode the pixel-wise depth distribution more accurately  (Table~\ref{tab:ablation_iteration_num}).

\customparagraph{Depth Prediction}
The depth map at iteration $k$ is predicted from the hidden state $h_k$ ($k=0, \cdots, K$). For simplicity, we omit index $k$ here. 
Recall that we let the hidden state encode the pixel-wise probability distribution of the depth. 
We extract those probabilities $\mathbf{P}~\in~\mathbb{R}^{W/4 \times H/4 \times D_2}$, for depths sampled uniformly at $D_2$ locations over the \textit{inverse} depth range, by applying 2D CNN on the hidden state, followed by a \textit{softmax} nonlinearity along the depth dimension. 

Usual strategies to predict a depth value from such sampled distributions are to take the \textit{argmax}~\cite{yao_2019_rmvsnet, d2hcrmvsnet} or the 
\textit{soft argmax}~\cite{kendall_2017_gcnet, yao_2018_mvsnet, gu_2020_cascademvsnet, patchmatchnet_wang}. 
The former corresponds to measuring the Kullback-Leibler divergence between a one-hot encoding of the ground truth and $\mathbf{P}$, but cannot deliver solutions beyond the discretization level (\eg `sub-pixel' solutions). %
The latter corresponds to measuring the distance of the expectation of $\mathbf{P}$ to the ground truth depth. 
While the expectation can take any continuous value, the measure cannot handle multiple modes in $\mathbf{P}$ and strictly prefers unimodal distributions.
At this point we propose a new hybrid strategy that combines classification and regression. It is robust to multi-modal distributions but also achieves `sub-pixel' precision that is not limited by the sampling resolution. 
Specifically, we find the index $\mathbf{X}(\mathbf{p})$ with the highest probability for pixel $\mathbf{p}$ from probability $\mathbf{P}$:
\begin{equation}
\setlength{\abovedisplayskip}{3pt}
\setlength{\belowdisplayskip}{3pt}
    \mathbf{X}(\mathbf{p}) = \argmax_j  \mathbf{P}(\mathbf{p},j).
\end{equation}
With a radius $r$, we take the expectation in the \textit{local inverse} range to compute the depth estimate $\mathbf{D}(\mathbf{p})$:
\begin{equation}
\setlength{\abovedisplayskip}{3pt}
\setlength{\belowdisplayskip}{3pt}
    \mathbf{D}(\mathbf{p}) \!\!=\!\! \left (\!\! \frac{1}{\sum_{j= \mathbf{X}(\mathbf{p})-r}^{\mathbf{X}(\mathbf{p})+r} \mathbf{P}(\mathbf{p},j)} \!\sum_{j= \mathbf{X}(\mathbf{p})-r}^{\mathbf{X}(\mathbf{p})+r} \!\!\frac{1}{d_j} \!\cdot\! \mathbf{P}(\mathbf{p},j)\!\!\right)^{\!\!\!-1}\!\!\!\!\!,
\end{equation}
where $d_j$ is the $j$-th depth value. 

\customparagraph{Confidence Estimation}
Because the hidden state of GRU models the pixel-wise depth probability distributions, we can estimate the uncertainty. 
We apply a 2D CNN followed by a \textit{sigmoid} on the hidden state $h$ to predict the confidence $\mathbf{C} \!\in\! \mathbb{R}^{W/4 \times H/4}$. 
The confidence is defined as the likelihood that the ground truth depth is located within a small range near the estimation~\cite{poggi2016learning,yao_2018_mvsnet,fu2017stereo,tosi2018beyond} 
and, thus, is used to filter outliers during the reconstruction~\cite{yao_2018_mvsnet} with a threshold $\tau$.

\subsection{Spatial Upsampling}
We upsample the depth map $\mathbf{D}_K$ output by the GRU probability estimator at the final iteration from $1/4$ to full resolution $\mathbf{D}_{\textrm{upsample}}$. 
The procedure is guided by image features: Given $\mathbf{F}_{0,2}$, the features of the reference, a 2D CNN predicts a mask $\mathbf{M} \!\in\! \mathbb{R}^{W/4 \times H/4 \times (4\times 4 \times 9)}$, where the last dimension represents the weights for the 9 nearest neighbors at coarse resolution. 
The depth at full resolution is then computed as the normalized (using \textit{softmax}) weighted combination of those neighbors~\cite{teed2020raft} (see supplementary). 
The confidence map is bilinearly upsampled to full resolution.

\subsection{Loss Function}
The loss function considers the depth and confidence estimated from the GRU, and the final upsampled depth. 
To facilitate convergence, we further utilize the coarse depth $\mathbf{D}_\textrm{initial} \in \mathbb{R}^{W/8 \times H/8}$ predicted from the similarity $\mathbf{\bar{S}}_\textrm{initial}$ in the Initialization phase. 
It is found as:
\begin{equation}
\setlength{\abovedisplayskip}{1pt}
\setlength{\belowdisplayskip}{1pt}
    \mathbf{D}_\textrm{initial}(\mathbf{p}) = \left (  \sum_{j= 1}^{D_1} \frac{1}{d_j} \cdot \mathbf{P}_\textrm{initial}(\mathbf{p},j)\right)^{-1},
\end{equation}
where the probability $\mathbf{P}_\textrm{initial}$ is generated from $\mathbf{\bar{S}}_\textrm{initial}$ by applying \textit{softmax} along the depth dimension. 
Then we bilinearly upsample $\mathbf{D}_\textrm{initial}$ to $1/4$ resolution.
The losses are based on %
$\mathbf{D}'$ that is converted from a depth map $\mathbf{D}$:
\begin{equation}\label{eq:normalize_loss}
\mathbf{D}'(\mathbf{p}) = \left(\frac{1}{\mathbf{D}(\mathbf{p})}-\frac{1}{d_{max}}\right)/\left(\frac{1}{d_{min}}-\frac{1}{d_{max}}\right).    
\end{equation}
We can summarize our loss function as follows:
\begin{equation}
\begin{aligned}
L_\textrm{full} = & \alpha^{K+1} L_\textrm{initial} + L_\textrm{upsample}\\
+ & \sum_{k=0}^{K} \alpha^{K-k} (L_{\textrm{class},k}+L_{\textrm{regress},k}+L_{\textrm{conf},k})
\end{aligned}
\raisetag{600pt}
\end{equation}
and only consider $N_\textrm{valid}$ pixels with valid ground truth depth.
The loss function is composed of five component-level losses: classification loss $L_{\textrm{class},k}$, regression losses $L_\textrm{initial}$, $L_{\textrm{regress},k}$, $L_\textrm{upsample}$ and confidence loss $L_{\textrm{conf},k}$,
\begin{equation}
\begin{split}
L_\textrm{initial} = & \;\beta\!\cdot\!||\mathbf{D}'_\textrm{initial}\minus \mathbf{D}'_{\textrm{gt},2}||_1, \\
L_{\textrm{class},k} = & \frac{1}{N_\textrm{valid}} \sum_{\mathbf{p}} \sum_{j=1}^{D_2} \minus\mathbf{Q}(\mathbf{p},j) \log (\mathbf{P}_k(\mathbf{p},j)), \\
L_{\textrm{regress},k} = & \;\beta\!\cdot|| \mathbb{I}\{|\mathbf{X}_{\textrm{gt},2} \minus \mathbf{X}_{k}| \leq r\}\!\cdot\!(\mathbf{D}'_{k}\minus\mathbf{D}'_{\textrm{gt},2})||_1, \\
L_{\textrm{conf},k} = & \minus \frac{1}{N_\textrm{valid}} \sum_{\mathbf{p}} [\mathbf{C}^*_k(p) \log \mathbf{C}_k(p) \\
& + (1\minus\mathbf{C}^*_k(p)) \log (1\minus\mathbf{C}_k(p))], \\
L_\textrm{upsample} = & \;\beta\!\cdot\! ||\mathbf{D}'_\textrm{upsample}\minus\mathbf{D}'_{\textrm{gt},0}||_1, \raisetag{.5cm}
\end{split}
\end{equation}
where $||\cdot||_1$ is the $l_1$ loss, $\mathbf{D}'_{\textrm{gt},l}$ is computed from the ground truth depth at level $l$ with Eq.~\ref{eq:normalize_loss}, 
$\mathbb{I}\{\cdot\}$ denotes the indicator function and $\alpha=0.8$ and $\beta=D_2$ are weights. 
For the confidence loss $L_{\textrm{conf},k}$, we set $\mathbf{C}^*_k$ to:
\begin{equation}
    \mathbf{C}^*_k(\mathbf{p}) = \mathbb{I}\{|\mathbf{D}'_{\textrm{gt},2}(\mathbf{p}) - \mathbf{D}'_k(\mathbf{p})| \leq \gamma\},
\end{equation}
where $\gamma$ is empirically set to $0.002$.
To train the classification, we define 
$\mathbf{Q}$ as the one-hot encoding generated by the ground truth depth $\mathbf{D}_{\textrm{gt},2}$, peaking at the nearest discrete location 
and use cross-entropy loss~\cite{yao_2019_rmvsnet} for $L_\textrm{class}$. 
The subsequent regression can only change the initial estimate of the classification by $r$ in any direction. 
Hence, the loss $L_{\textrm{regress}}$ only considers pixels whose ground truth classification index, 
$\mathbf{X}_{\textrm{gt},2}(\mathbf{p})$, falls within a radius $r$ of the estimated index  $\mathbf{X}_{k}(\mathbf{p})$. 
For the $1^{st}$ epoch of the training, we exclude $L_\textrm{regress}$ and $L_\textrm{conf}$ from the loss to warm up the classification.

%% file: tex/experiments.tex
\section{Experiments}

\subsection{Datasets}
The DTU dataset~\cite{aanaes2016_dtu} is an indoor multi-view stereo dataset with 124 different scenes and 7 different lighting conditions. 
We use the training, testing and validation split introduced in~\cite{ji_2017_surfacenet}.
BlendedMVS~\cite{yao2020blendedmvs} is a large-scale dataset, which provides over 17k high-quality training samples of various scenes. 
Tanks \& Temples~\cite{knapitsch2017tanks} is a large-scale outdoor dataset consisting of complex environments that are captured under real-world conditions.
The ETH3D benchmark~\cite{2017eth3d} consists of calibrated high-resolution images of real-world scenes with strong viewpoint variations. %

\subsection{Implementation Details}
Implemented with PyTorch, two models are trained on DTU~\cite{aanaes2016_dtu} (\textbf{Ours}) and \textbf{L}arge-\textbf{S}cale BlendedMVS~\cite{yao2020blendedmvs} (\textbf{Ours-LS}) respectively. 
For DTU, we use an image resolution of $640 \times 512$ and the number of input images to $N = 5$. 
Observing that a fixed depth range (\eg MVSNet~\cite{yao_2018_mvsnet}) filters out some of the ground truth signals on DTU, 
we propose to estimate a depth range per-view from the sparse ground truth point clouds for both training and evaluation. 
This is comparable to the treatment of real-world scenes~\cite{knapitsch2017tanks, 2017eth3d} in MVS, 
where the per-view depth range is usually estimated from a sparse SfM reconstruction, \eg~\cite{schoenberger2016colmap, moulon2016openmvg}.
For BlendedMVS, we use an image resolution of $768 \times 576$ and $N = 5$ input images. Here, the depth range is provided by the dataset. 
To improve the robustness during training, we follow~\cite{patchmatchnet_wang} to randomly select the source views and also scale the scene within the range of $[0.8,1.25]$. 

We use $D_1\!=\!32$ during initialization and perform $K=4$ update iterations. 
Here, we set $R_1=2^{-7}, R_2=2^{-5}, R_3=2^{-3}$ and $N_1=4, N_2=4, N_3=2$. 
When predicting the depth, we use $D_2=256$ and $r=4$. 
We train for 16 epochs using Adam~\cite{kingma2015adam} ($\beta_1 \!=\! 0.9, \beta_2 \!=\! 0.999$). 
The learning rate is initially set to 0.001, and halved after 4, 8 and 12 epochs. 
We set the batch size to 4 and 2 for the models trained on DTU and BlendedMVS. The models are trained on a single Nvidia RTX 2080Ti GPU. 
After depth estimation, we perform confidence filtering (threshold $\tau$ is set to $0.3$) and geometric consistency filtering to remove outliers and then reconstruct the point clouds (see supplementary).

\subsection{Benchmark Performance}

\begin{table}
\setlength{\belowcaptionskip}{-0.5cm}
\setlength{\abovecaptionskip}{0.1cm}
\centering
\footnotesize
\begin{widetable}{\columnwidth}{cccc}
 \hline
 Methods & Acc.(mm) $\downarrow$ & Comp.(mm) $\downarrow$ & Overall(mm) $\downarrow$\\
 \hline
 Camp~\cite{camp} & 0.835 & 0.554 & 0.695 \\ 
 Furu~\cite{furukawa2010} & 0.613 & 0.941 & 0.777 \\
 Tola~\cite{tola2012efficient} & 0.342 & 1.190 & 0.766 \\
 Gipuma~\cite{galliani_2015_gipuma} & \textbf{0.283} & 0.873 & 0.578 \\
 \hline
 MVSNet~\cite{yao_2018_mvsnet} & 0.396 & 0.527 & 0.462\\
 R-MVSNet~\cite{yao_2019_rmvsnet} & 0.383 & 0.452 & 0.417\\
 CIDER~\cite{xu2020learning_inverse} & 0.417 & 0.437 & 0.427\\
 P-MVSNet~\cite{luo_2019_p_mvsnet} & 0.406 & 0.434 & 0.420\\
 Point-MVSNet~\cite{chen_2019_pointmvsnet} & 0.342 & 0.411 & 0.376\\
 Fast-MVSNet~\cite{yu_2020_fastmvsnet} & 0.336 & 0.403 & 0.370\\ 
 CasMVSNet~\cite{gu_2020_cascademvsnet} & 0.325 & 0.385 & 0.355\\
 UCS-Net~\cite{cheng_2020_ucsnet} & 0.338 & 0.349 & \textbf{0.344} \\
 CVP-MVSNet~\cite{yang_2020_cvpr} & 0.296 & 0.406 & 0.351 \\
 Vis-MVSNet~\cite{vismvsnet} & 0.369 & 0.361 & 0.365 \\
 $D^2$HC-RMVSNet~\cite{d2hcrmvsnet} & 0.395 & 0.378 & 0.386 \\
 PatchmatchNet~\cite{patchmatchnet_wang} & 0.427 & \textbf{0.277} & 0.352\\
 Ours & 0.373 & 0.354 & 0.363 \\
 \hline
\end{widetable}
\caption{Quantitative results of different methods on DTU~\cite{aanaes2016_dtu}.
Methods are separated into two categories (from top to bottom): traditional and trained on DTU~\cite{aanaes2016_dtu}
}
\label{tab:evaluation_dtu}
\end{table}
\customparagraph{Evaluation on DTU}
We use the DTU-only trained model for evaluation. 
Image size, number of views $N$ and number of iterations $K$ are set to $1600 \!\times\! 1152$, 5 and 4.
In our quantitative evaluation, we use the metrics provided by DTU~\cite{aanaes2016_dtu} and calculate \textit{accuracy}, \textit{completeness} and \textit{overall quality} as the mean of both metrics~\cite{yao_2018_mvsnet}.
Shown in Table~\ref{tab:evaluation_dtu}, 
our method achieves competitive performance in \textit{overall quality}. %
Compared to the highly efficient PatchmatchNet~\cite{patchmatchnet_wang}, our method achieves higher \textit{accuracy}.

\begin{figure}
\centering
\setlength{\belowcaptionskip}{-0.6cm}
\setlength{\abovecaptionskip}{0.1cm}
{\includegraphics[width=0.85\linewidth]{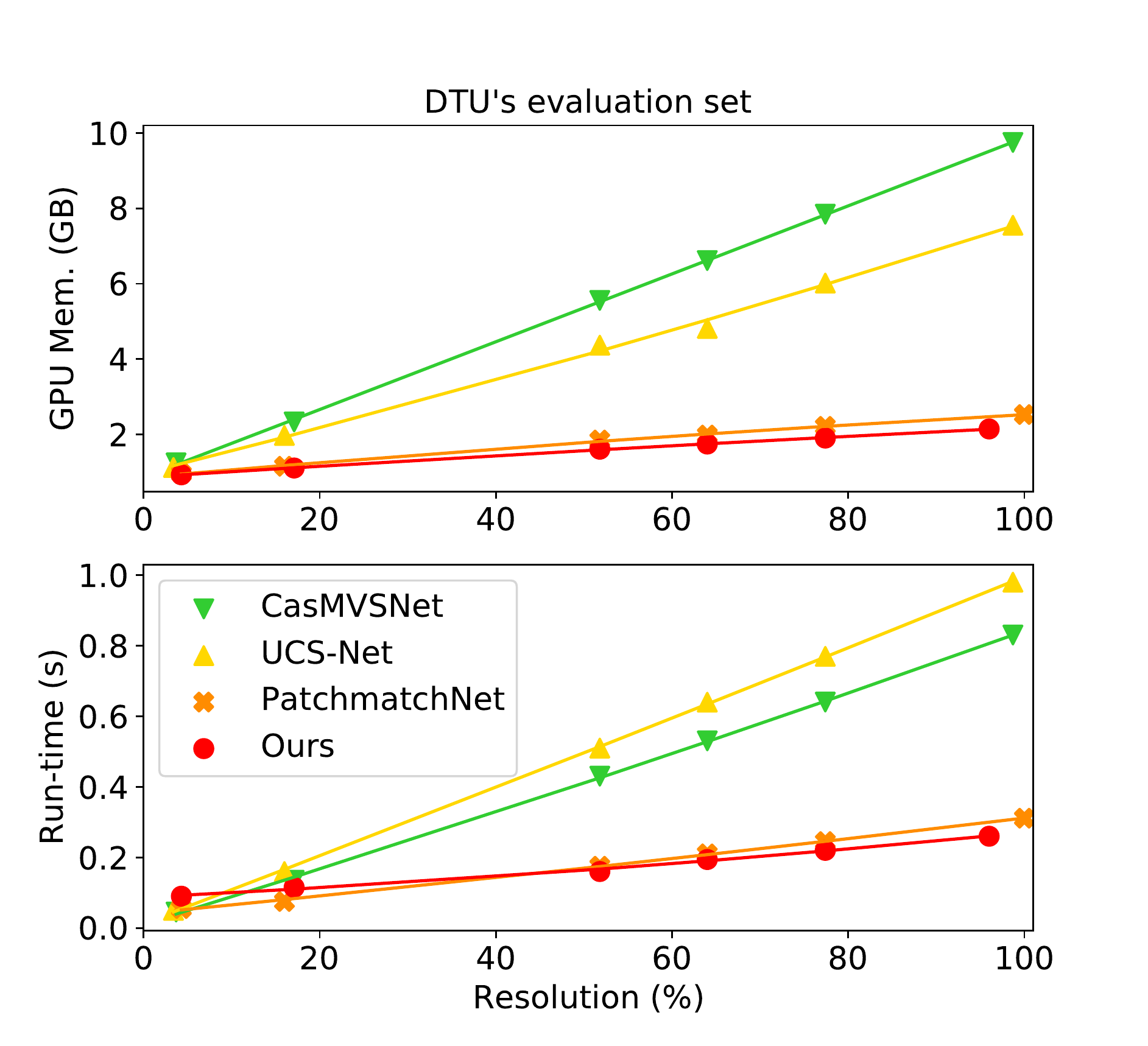}}
\caption{
    Relating GPU memory and run-time  to the image resolution on DTU's evaluation set~\cite{aanaes2016_dtu}. 
    The original image resolution is $1600\!\times\!1200$ (100\%). 
    }
\label{fig:memory_time_compare}
\end{figure}

\customparagraph{Memory and Run-time Comparison}
To demonstrate the high efficiency of our method, we compare to the learning-based multi-stage methods~\cite{gu_2020_cascademvsnet, cheng_2020_ucsnet, patchmatchnet_wang} that are tailored for efficiency in both memory and run-time. 
The experiments use the same environment as before and are summarized in Fig.~\ref{fig:memory_time_compare}. 
For our method, both memory consumption and run-time increase the slowest 
with the input size.
For example, at a resolution of $1600 \!\times\! 1152$, 
memory consumption and run-time are reduced by 78.0\% and 68.7\% compared to CasMVSNet~\cite{gu_2020_cascademvsnet}, 
by 71.6\% and 73.5\% compared to UCS-Net~\cite{cheng_2020_ucsnet} and 
by 15.2\% and 16.1\% compared to PatchmatchNet~\cite{patchmatchnet_wang}, the most efficient learning-based MVS method so far. 
Combined with the quantitative analysis in Table~\ref{tab:evaluation_dtu}, we conclude that our method demonstrates higher efficiency in both memory and run-time than most state-of-the-art learning-based methods, at a competitive performance.

\customparagraph{Evaluation on Tanks \& Temples}
We set the input image size to $1920 \times 1024$, the number of views, $N$, to 7 and the number of iterations, $K$, to 4.
The camera parameters and depth ranges are estimated with OpenMVG~\cite{moulon2016openmvg}. 
During evaluation, the GPU memory consumption and run-time for estimating each depth map are 2426~MB and 0.405~s respectively. 
Among the methods trained on DTU~\cite{aanaes2016_dtu}, as shown in Table~\ref{tab:evaluation_tank}, the \textit{F-score} of our method on the intermediate dataset is similar to CasMVSNet~\cite{gu_2020_cascademvsnet} and much better than many other multi-stage methods~\cite{cheng_2020_ucsnet, patchmatchnet_wang, yang_2020_cvpr}.
On the complex advanced dataset, our method performs best in \textit{Recall} and \textit{F-score} among all the methods. 
Compared with PatchmatchNet~\cite{patchmatchnet_wang}, our method performs much better on all the metrics of both datasets.
Training on the large-scale BlendedMVS~\cite{yao2020blendedmvs} dataset appears to improve the generalization ability of our method on both datasets compared with the model trained on DTU only. 
On the advanced dataset, our method performs better in \textit{F-score} than Vis-MVSNet~\cite{vismvsnet}, which is a state-of-the-art multi-stage method. 
Overall, operating at a notably low memory and run-time, our method demonstrates very competitive generalization performance compared to most learning-based methods.

\begin{table*}
\setlength{\belowcaptionskip}{-0.3cm}
\setlength{\abovecaptionskip}{0.1cm}
\centering
\footnotesize
\begin{tabular}{c|ccc|ccc}
 \hline
\multirow{2}{*}{Methods} & \multicolumn{3}{c|}{Intermediate Dataset}  & \multicolumn{3}{c}{Advanced Dataset}\\
 \cline{2-7}
  & Precision $\uparrow$ & Recall $\uparrow$ & F-score $\uparrow$ & Precision $\uparrow$ & Recall $\uparrow$ & F-score $\uparrow$ \\
 \hline
 COLMAP~\cite{schoenberger2016colmap} & 43.16 & 44.48 & 42.14  & \textbf{33.65} & 23.96 & 27.24 \\ 
 \hline
 MVSNet~\cite{yao_2018_mvsnet} & 40.23 & 49.70 & 43.48 &- &- & - \\
 R-MVSNet~\cite{yao_2019_rmvsnet} & 43.74 & 57.60 & 48.40 &31.47 & 22.05 & 24.91 \\
 CIDER~\cite{xu2020learning_inverse} & 42.79 & 55.21 & 46.76 & 26.64 & 21.27 & 23.12 \\
 P-MVSNet~\cite{luo_2019_p_mvsnet} & 49.93 & 63.82 & 55.62 &- &- & - \\
 Point-MVSNet~\cite{chen_2019_pointmvsnet} & 41.27 & 60.13 & 48.27 &- &- & - \\
 Fast-MVSNet~\cite{yu_2020_fastmvsnet} & 39.98 & 59.89 & 47.39 &- &- & - \\
 CasMVSNet~\cite{gu_2020_cascademvsnet} & 47.62 & 74.01 & 56.84 & 29.68 & 35.24 & 31.12 \\
 UCS-Net~\cite{cheng_2020_ucsnet} & 46.66 & 70.34 & 54.83 &- &- & - \\
 CVP-MVSNet~\cite{yang_2020_cvpr} & 51.41 & 60.19 & 54.03 &- &- & - \\
 $D^2$HC-RMVSNet~\cite{d2hcrmvsnet} & 49.88 & 74.08 & 59.20 &- &- & - \\
 PatchmatchNet~\cite{patchmatchnet_wang} & 43.64 & 69.37 & 53.15 & 27.27 & 41.66 & 32.31 \\
 Ours & 46.82 & 73.50 & 56.22 & 28.04 & 42.60 & 33.24 \\
 \hline
 PatchMatch-RL~\cite{lee2021patchmatch} & 45.91 & 62.30 & 51.81 & 30.57 & 36.73 & 31.78\\
 Vis-MVSNet~\cite{vismvsnet} & \textbf{54.44} & 70.48 & \textbf{60.03}  & 30.16 & 41.42 & 33.78\\
 Ours-LS & 47.53 & \textbf{74.69} & 56.94 & 28.70 & \textbf{44.19} & \textbf{34.17} \\
 \hline

\end{tabular}

\caption{Results of different methods on Tanks \& Temples~\cite{knapitsch2017tanks}. Methods are separated into three categories (from top to bottom): traditional, trained on DTU~\cite{aanaes2016_dtu} and trained on BlendedMVS~\cite{yao2020blendedmvs}.
}
\label{tab:evaluation_tank}
\end{table*}

\customparagraph{Evaluation on ETH3D Benchmark}
We set the input image size to $1920 \!\times\! 1280$, the number of views $N$ to 7 and the number of iterations $K$ to 4.
The camera parameters and depth ranges are estimated with COLMAP~\cite{schoenberger2016colmap}.
During evaluation, the GPU memory consumption and run-time for estimating each depth map are 2898~MB and 0.410~s respectively. 
The results are summarized in Table~\ref{tab:evaluation_eth}.
When trained on DTU~\cite{aanaes2016_dtu}, our method performs much better in \textit{accuracy} than PVSNet~\cite{xu2020pvsnetpv} and PatchmatchNet~\cite{patchmatchnet_wang}. On the training dataset, our method achieves comparable performance in \textit{$F_1$-score} as PVSNet~\cite{xu2020pvsnetpv} and COLMAP~\cite{schoenberger2016colmap}. On the test dataset, the performance of our method in \textit{$F_1$-score} is better than PatchmatchNet~\cite{patchmatchnet_wang}, PVSNet~\cite{xu2020pvsnetpv} and COLMAP~\cite{schoenberger2016colmap}, which is competitive to ACMH~\cite{xu_2019_acmm}. 
When trained on BlendedMVS~\cite{yao2020blendedmvs}, the performance of our method improves significantly in all metrics on both datasets compared with the model solely trained on DTU. Our method performs best in \textit{$F_1$-score} on both datasets among all competitors. 
This analysis further demonstrates the effectiveness, efficiency and generalization capabilities of our method.

\begin{table*}[ht]
\setlength{\belowcaptionskip}{-0.4cm}
\setlength{\abovecaptionskip}{0.1cm}
\centering
\footnotesize
\begin{tabular}{c|ccc|ccc}
 \hline
\multirow{2}{*}{Methods} & \multicolumn{3}{c|}{Training Dataset}  & \multicolumn{3}{c}{Test Dataset}\\
 \cline{2-7}
  & Accuracy $\uparrow$ & Completeness $\uparrow$ & $F_1$-score $\uparrow$ & Accuracy $\uparrow$ & Completeness $\uparrow$ & $F_1$-score $\uparrow$ \\
 \hline
 Gipuma~\cite{galliani_2015_gipuma} & 84.44 & 34.91 & 36.38 & 86.47 & 24.91  & 45.18 \\ 
 PMVS~\cite{furukawa2010} & 90.23 & 32.08 & 46.06 & 90.08 & 31.84 & 44.16 \\
 COLMAP~\cite{schoenberger2016colmap} & \textbf{91.85} & 55.13 & 67.66  & \textbf{91.97} & 62.98 & 73.01  \\
 ACMH~\cite{xu_2019_acmm} & 88.94 & 61.59 & 70.71 & 89.34 & 68.62 & 75.89\\
 \hline
 PVSNet~\cite{xu2020pvsnetpv} & 67.84 & 69.66 & 67.48 & 66.41 & 80.05 & 72.08  \\ 
 PatchmatchNet~\cite{patchmatchnet_wang} & 64.81 & 65.43 & 64.21 & 69.71 & \textbf{77.46} & 73.12  \\
 Ours & 73.62 & 61.87 & 66.36 & 76.91 & 72.65 & 74.29  \\
 \hline
 PatchMatch-RL~\cite{lee2021patchmatch} & 76.05 & 62.22 & 67.78 & 74.48 & 72.06 & 72.38\\
 Ours-LS & 79.79 & \textbf{66.08} & \textbf{71.69} & 84.73 & 76.49 & \textbf{80.06}  \\
 \hline

\end{tabular}

\caption{Results of different methods on ETH3D~\cite{2017eth3d} (evaluation threshold is 2cm). Methods are separated into three categories (from top to bottom): traditional, trained on DTU~\cite{aanaes2016_dtu} and trained on BlendedMVS~\cite{yao2020blendedmvs}. 
}
\label{tab:evaluation_eth}
\end{table*}

\subsection{Ablation Study}
We conduct an ablation study of our model trained on DTU~\cite{aanaes2016_dtu} 
to analyze the effectiveness of its components. 
Results of the first six experiments are summarized in Table~\ref{tab:ablations}.

\begin{table*}
\setlength{\belowcaptionskip}{-0.375cm}
\setlength{\abovecaptionskip}{0.1cm}
\centering
\footnotesize
\begin{tabular}{c|c|ccc|c}
 \hline
\multirow{2}{*}{Experiments} & \multirow{2}{*}{Methods} & \multicolumn{3}{c|}{DTU dataset}  & ETH3D Training\\
 \cline{3-6}
  & & Acc.(mm) $\downarrow$ & Comp.(mm) $\downarrow$ & Overall(mm) $\downarrow$ & $F_1$-score $\uparrow$ \\
 \hline
 \multirow{2}{*}{Depth Prediction} & \underline{Classification+Regression} & 0.373 & 0.354 & 0.363 & 66.36  \\
  & Classification Only & 0.448 & 0.408 & 0.428 & 63.34\\
  & Regression Only & 0.400 & 0.415 & 0.408 & 50.19\\
 \hline
 \multirow{2}{*}{Confidence} & \underline{Learned} & 0.373 & 0.354 & 0.363 & 66.36  \\
 & Sum of Probabilities & 0.374 & 0.355 & 0.364 & 66.04\\
 \hline
 \multirow{2}{*}{Scale of Feature} & \underline{Multi-scale} & 0.373 & 0.354 & 0.363 & 66.36  \\
 & Single-scale & 0.381 & 0.359 & 0.370 & 64.02 \\
 \hline
 \multirow{2}{*}{Depth Upsampling} & \underline{Learned} & 0.373 & 0.354 & 0.363 & 66.36  \\
 & Bilinear & 0.409 & 0.366 & 0.387 & 63.79\\
  \hline
 \multirow{2}{*}{Inverse Depth Loss} & \underline{Yes} & 0.373 & 0.354 & 0.363 & 66.36  \\
 & No & 0.379 & 0.349 & 0.364 & 64.51\\
 \hline
 \multirow{2}{*}{Pixel-wise View Weight} & \underline{Yes} & 0.373 & 0.354 & 0.363 & 66.36  \\
 & No & 0.371 & 0.368 & 0.369 & 60.02\\
 
 \hline

\end{tabular}

\caption{Ablation study of the first six experiments on DTU~\cite{aanaes2016_dtu} and ETH3D~\cite{2017eth3d}. Settings used in our method are underlined. 
}
\label{tab:ablations}
\end{table*}

\customparagraph{Depth Prediction}
For the depth estimated from GRU, instead of using both classification and regression, we also try both \textit{argmax} (classification only) and \textit{soft argmax} (regression only), \ie taking the expectation on all the $D_2$ depth samples. 
Our hybrid strategy performs better on both DTU~\cite{aanaes2016_dtu} and ETH3D~\cite{2017eth3d}. 
Apparently, the gap on the unseen ETH3D data widens when (also) utilizing the classification loss, 
whereas classification only leads to inferior accuracy on DTU. 
We conjecture that \textit{soft argmax} (regression) only training could be more prone to overfitting, 
implied by the tendency to force single peaked distributions.

\begin{figure}[htbp]
\centering
\setlength{\abovecaptionskip}{-0.15cm}
\setlength{\belowcaptionskip}{-0.4cm}
{\includegraphics[width=1.0\linewidth]{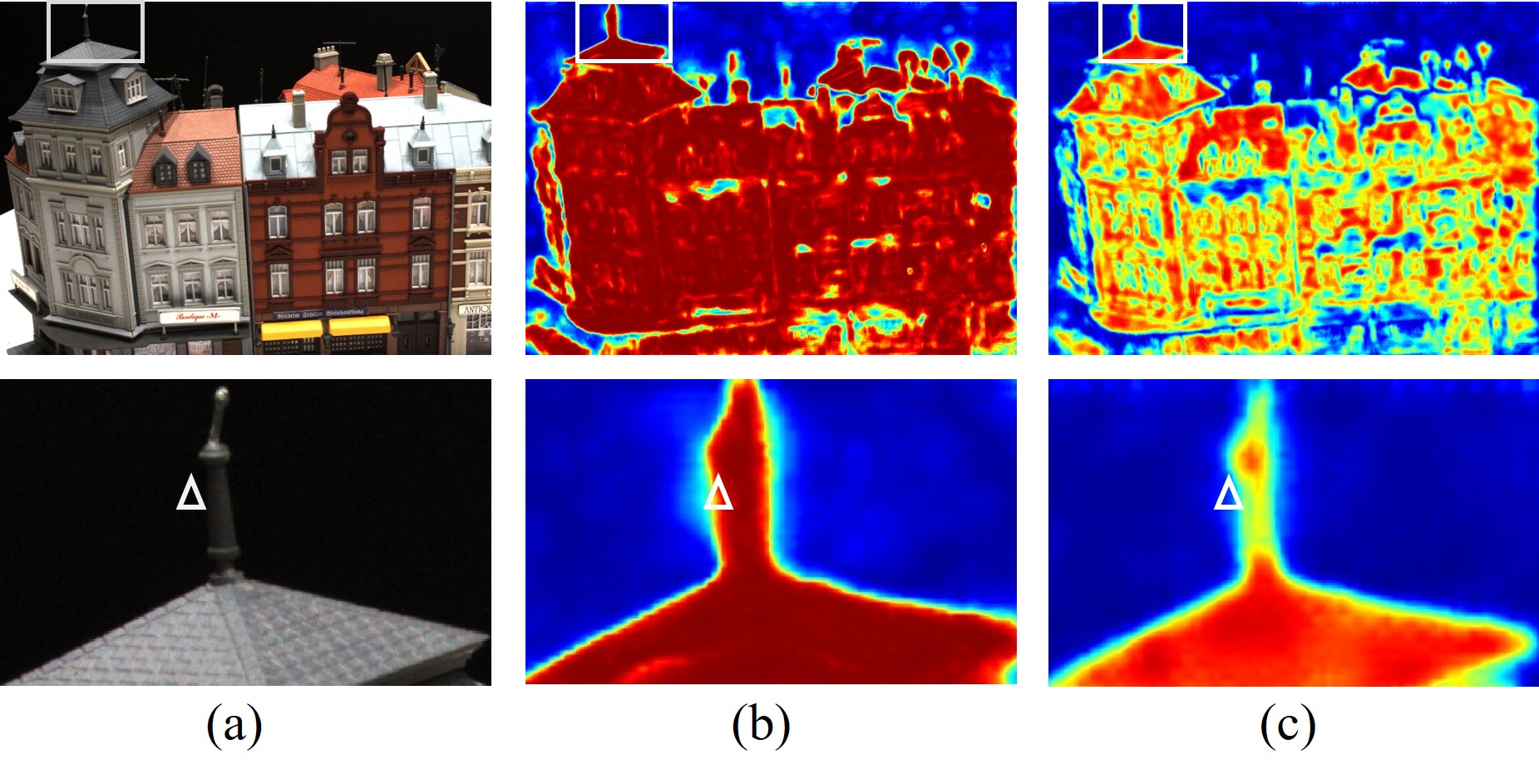}}
\caption{
Visualization of confidence. (a) Reference image (chosen background pixel near the object boundary is highlighted). (b) Confidence as sum of probabilities. (c) Our learned confidence. Note that our method correctly outputs a lower confidence for the background pixel compared to the foreground pixels.
}
\label{fig:abla_confidence}
\end{figure}

\customparagraph{Confidence}
We compare our idea of explicitly learning the confidence from the hidden state to a commonly used strategy that defines the confidence 
as the sum of 4 probability samples near the estimate~\cite{yao_2018_mvsnet} from the probability volume $\mathbf{P}$. 
Following MVSNet~\cite{yao_2018_mvsnet}, we use a threshold of $\tau\!=\!0.8$ for 3D reconstruction. 
Our strategy performs slightly better on DTU~\cite{aanaes2016_dtu} and ETH3D~\cite{2017eth3d}.
A visualization in Fig.~\ref{fig:abla_confidence} shows that our learned confidence can deliver more accurate predictions at object boundaries. 
When taking confidence as a sum of probabilities, the un-matchable uniformly colored background pixel near the boundary receives a confidence as high as  foreground pixels, due to an over-smoothing effect induced by averaging  probabilities.

\customparagraph{Scale of Feature}
Recall that we utilize multi-scale features for similarity computation throughout the Iterative Update. 
Here, we also try to only use level 2 features (we still employ level 3 features in the Initialization phase to reduce computation).
Our analysis shows that our model benefits from using multi-scale context information on both DTU~\cite{aanaes2016_dtu} and ETH3D~\cite{2017eth3d}.

\customparagraph{Depth Upsampling}
We compare learned upsampling to simple bilinear upsampling. Clearly, the learned approach delivers better results on both DTU~\cite{aanaes2016_dtu} and ETH3D~\cite{2017eth3d}.

\begin{table}
\setlength{\belowcaptionskip}{-0.4cm}
\setlength{\abovecaptionskip}{0.1cm}
\centering
\footnotesize
 \begin{tabular}{ccccc}
 \hline
 $K$ & Acc.(mm) $\downarrow$ & Comp.(mm) $\downarrow$ & Overall(mm) $\downarrow$ & Time (s)\\
 \hline
 1 & 0.389 & 0.403 & 0.396 & \textbf{0.190}\\ 
 2 & 0.383 & 0.381 & 0.382 & 0.220 \\
 3 & 0.377 & 0.363 & 0.370 & 0.250 \\
 4 & 0.373 & 0.354 & 0.363 & 0.260 \\
 8 & 0.366 & 0.339 & 0.353 & 0.375 \\
 16 & \textbf{0.363} & \textbf{0.333} & \textbf{0.348} & 0.610 \\
 \hline
\end{tabular}

\caption{
    Ablation study on the number of GRU iterations $K$ on DTU~\cite{aanaes2016_dtu}. }
\label{tab:ablation_iteration_num}
\end{table}

\customparagraph{Inverse Depth Loss} 
By default, we compute the losses $L_\textrm{initial}$, $L_\textrm{regress}$, $L_\textrm{upsample}$ \wrt the \textit{inverse} depth. We also try computing the losses without converting to the \textit{inverse} depth range. 
While the result on DTU~\cite{aanaes2016_dtu} is similar, the performance on the large-scale ETH3D~\cite{2017eth3d}  dataset, is explicitly improved when using inverse depth.

\customparagraph{Pixel-wise View Weight}
By default, we estimate the pixel-wise view weight to robustly integrate the matching information from the source views. 
Here, we compare to the simpler approach of taking the average over the views. 
The large performance difference on ETH3D~\cite{2017eth3d} can be explained 
by strong viewpoint variations within the dataset, such that our robust 
integration of the matching information across the views becomes very impactful. 

\customparagraph{Number of Iterations}
Table~\ref{tab:ablation_iteration_num} relates the performance in \textit{accuracy}, \textit{completeness} and \textit{overall quality} to the number of iterations on the GRU. 
Running more iterations allows to inject more matching information into the hidden state, 
leading to more accurate depth probability distributions per pixel. 
Although we limit our model to $K\!=\!4$ iterations during training, the performance keeps improving beyond that in all metrics. 
Comparing the result with Table~\ref{tab:evaluation_dtu}, we can even observe that our method achieves state-of-the-art performance at $K\!=\!16$ on DTU. 
Despite the increased run-time, the inference is still faster than most multi-stage methods~\cite{gu_2020_cascademvsnet, cheng_2020_ucsnet, yang_2020_cvpr}. 
We conclude by noting that the flexibility to run our model for an arbitrary number of iterations 
enables the user to trade-off time efficiency for performance in dependence of the application.

\customparagraph{Number of Views}
We vary the number of views $N$ and summarize the results in Table~\ref{tab:ablation_num_views}.
Multi-view information can help to alleviate problems such as occlusions and the reconstruction quality improves at a higher value, saturating at around 6 views.

\begin{table}
\setlength{\belowcaptionskip}{-0.4cm}
\setlength{\abovecaptionskip}{0.1cm}
\centering
\footnotesize
 \begin{tabular}{cccc}
 \hline
 $N$ & Acc.(mm) $\downarrow$ & Comp.(mm) $\downarrow$ & Overall(mm) $\downarrow$ \\
 \hline
 2 & 0.376 & 0.556 & 0.466 \\ 
 3 & \textbf{0.370} & 0.387 & 0.379 \\
 4 & 0.372 & 0.360 & 0.366 \\
 5 & 0.373 & 0.354 & 0.363 \\
 6 & 0.373 & \textbf{0.352} & \textbf{0.362} \\
 7 & 0.371 & 0.356 & 0.364 \\
 \hline
\end{tabular}

\caption{
    Ablation study on the number of views $N$ on DTU~\cite{aanaes2016_dtu}. }
\label{tab:ablation_num_views}
\end{table}

\subsection{Limitations}
While our network structure allows to trade-off speed and accuracy by adjusting the number of iterations during inference, 
we need to fix the number of samples that our probability distributions consist of. 
This number, $D_2$, is afterwards determined by the network structure and cannot be adjusted for different scenes. 
Likewise predetermined is the range of the data term samples placed around the current solution
that are ingested into the CNN. In our model those samples cover $2R_3\!=\!1/4^\textrm{th}$ of the total \textit{inverse} depth range. 

%% file: tex/conclusion.tex
\section{Conclusion}
We present IterMVS, a novel learning-based MVS method combining highest efficiency and competitive reconstruction quality. %
We propose to explicitly encode a pixel-wise probability distribution of depth in the hidden state of a GRU-based estimator.
In each iteration, we inject multi-scale matching information and extract the -- in the \textit{inverse} depth range -- uniformly sampled depth distribution to estimate depth map and confidence.
Extensive experiments on DTU, Tanks \& Temples and ETH3D show highest efficiency in both memory and run-time, and a better generalization ability than many state-of-the-art learning-based methods.

%% file: tex/appendix.tex
\newpage
\begin{center}
      {\Large \bf Appendix}
\end{center}

\setcounter{section}{0}

\section{Matching Similarities with Multi-scale Features}
When performing the iterative updates, we compute the matching similarities from features on all levels to include multi-scale information. 
For a pixel $\mathbf{p}$ with coordinates $(x,y)$ in the depth-map $\mathbf{D} \in \mathbb{R}^{W/4 \times H/4}$, 
we first find its corresponding position $\mathbf{p}_l$ for level $l$ ($l=1,2,3$) in
the reference feature map at the coordinates $(x/2^{l-2},y/2^{l-2})$. 
Then, the reference feature of $\mathbf{p}_l$, $\mathbf{F}_{0,l}(\mathbf{p}_l)$, is found via bilinear interpolation. 
Afterwards, with $N_l$ new depth hypotheses, known camera parameters (for each level $l$)  and source features ${\left\{\mathbf{F}_{i,l}\right\}}_{i=1}^{N-1}$, 
we warp (using differentiable warping) $\mathbf{p}_l$ into the respective source view and compute the matching similarities between reference and each source view. 
Finally, we use the $2\times$ upsampled pixel-wise view weights to compute the integrated matching similarities 
and pass them through a level-wise 2D U-Net to aggregate the neighborhood information. 

\section{Depth Upsampling}
\begin{figure}[ht]
\centering
{\includegraphics[width=1.0\linewidth]{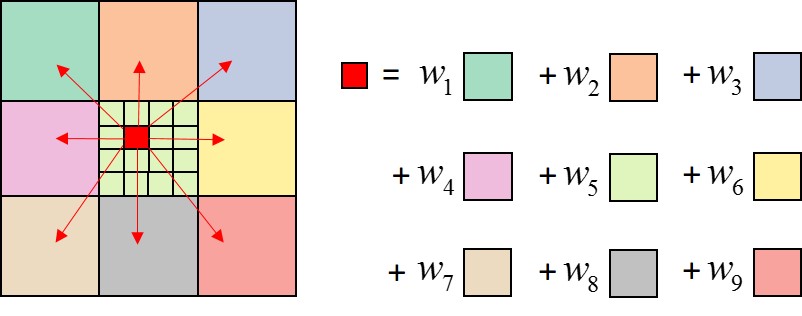}}
\caption{Illustration of depth upsampling. In the high resolution depth map, the depth of each pixel is the weighted sum of its 9 coarse resolution neighbors. }
\label{fig:upsampling}
\end{figure}

Following RAFT~\cite{teed2020raft}, we upsample the depth map from $1/4$ to full resolution.
Specifically, the depth of each pixel in the high resolution depth map is a convex combination of its 9 neighbors at the coarse resolution. 
The weights are learned from the reference feature map. 
Fig.~\ref{fig:upsampling} illustrates the upsampling process. 

\section{Point Cloud Reconstruction}
Before fusing the depth maps, we filter out unreliable depth estimates, following MVSNet~\cite{yao_2018_mvsnet}. 
There are two filtering steps: geometric consistency filtering and confidence filtering. 

\customparagraph{Geometric Consistency Filtering}
Following MVSNet~\cite{yao_2018_mvsnet}, we apply a geometric
constraint to measure the consistency of depth estimates among multiple views.
For each pixel $\mathbf{p}$ in the reference view, we project it, using its depth $d_0$, to a pixel $\mathbf{p}_i$ in the $i$-th source view. 
After looking up its depth $d_i$ in the source view, we reproject $\mathbf{p}_i$ into the reference view, 
and look up the depth $d_\textrm{reproj}$ at this location,  $\mathbf{p}_\textrm{reproj}$. 
We consider pixel $\mathbf{p}$ and its depth $d_0$ as consistent to the $i$-th source view,  
if the distances, in image space and depth, between the original estimate and its reprojection satisfy: 
\begin{equation}
    |\mathbf{p}_\textrm{reproj}-\mathbf{p}| < \delta, |d_\textrm{reproj}-d_0|/d_0 < \varepsilon,
\end{equation}
where $\delta=1$ and $\varepsilon=0.01$ are two thresholds. 
Finally, we accept the estimations as reliable, if they are consistent in at least $N_\textrm{geo}$ source views. 

\customparagraph{Confidence Filtering}
Since our learned confidence indicates how close the estimation is to the ground truth depth, we use it to filter out estimations with high uncertainty. 
Specifically, we use a confidence threshold $\tau=0.3$ throughout the experiments to filter out all the pixels with confidence lower than it. 

\section{Visualization of Probability}
\begin{figure*}[ht]
\centering
{\includegraphics[width=0.65\linewidth]{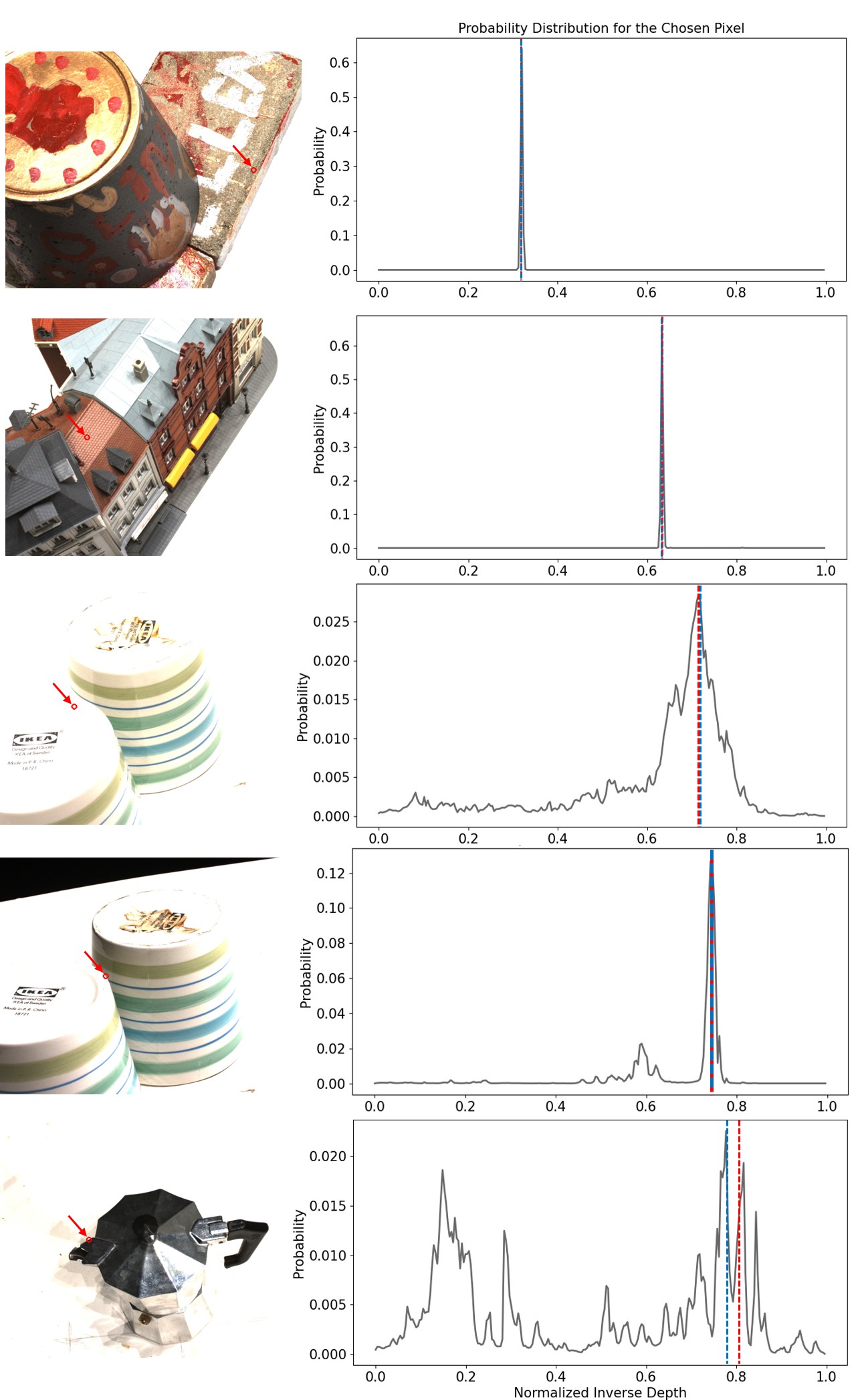}}
\caption{Visualization of probability. Left: Reference images (the chosen pixels are highlighted). Right: Probability distribution of depth for the chosen pixels. \textcolor{red}{Red line} denotes ground truth depth and \textcolor{blue}{blue line} denotes our estimation.}
\label{fig:visual_prob}
\end{figure*}

\begin{figure*}[ht]
\centering
{\includegraphics[width=0.9\linewidth]{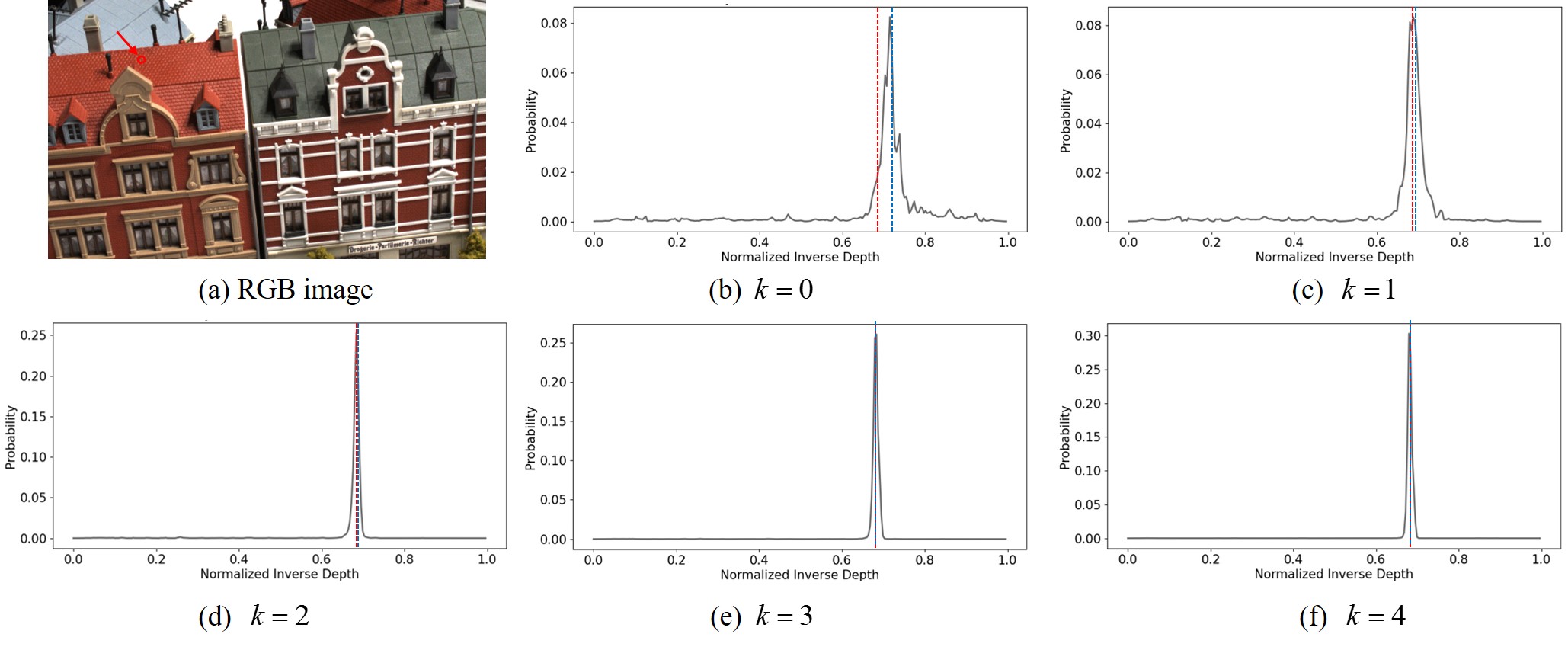}}
\caption{(a) Reference image (the chosen pixel is highlighted). (b)-(f) Probability distribution of depth for the chosen pixel in the $k$-th GRU iteration ($k=0$ represents the probability distribution from initial hidden state $h_0$). \textcolor{red}{Red line} denotes ground truth depth and \textcolor{blue}{blue line} denotes our estimation. Note our estimation becomes more accurate while several local maxima get suppressed in the distribution that converges to a single, more pronounced peak with more iterations.}
\label{fig:visual_prob_iter}
\end{figure*}

Our GRU-based probability estimator encodes the per-pixel probability distribution of depth with the hidden state. 
A 2D CNN is applied on the hidden state to estimate the probability of $D_2$ samples that are uniformly distributed in the \textit{inverse} depth range for each pixel. 
We visualize this probability distribution for various scenes in Fig.~\ref{fig:visual_prob}. 
For pixels with distinct features, the probability has a single dominant peak and the estimation is precise. 
For some challenging situations, where distributions are non-peaky or multi-modal, \eg in textureless areas, our hybrid depth estimation strategy can still robustly produce estimations as accurate as possible. 
We also visualize the update process of probability distribution in Fig.~\ref{fig:visual_prob_iter}. We observe that the probability distribution becomes more focused, several local maxima get suppressed and the estimation becomes more precise with more GRU iterations. In each iteration, multi-scale matching information is injected into the hidden state.  
This allows the hidden state to more accurately model the per-pixel probability distribution of depth with each iteration. 

\section{Visualization of Pixel-wise View Weight}
\begin{figure*}[ht]
\centering
{\includegraphics[width=1.0\linewidth]{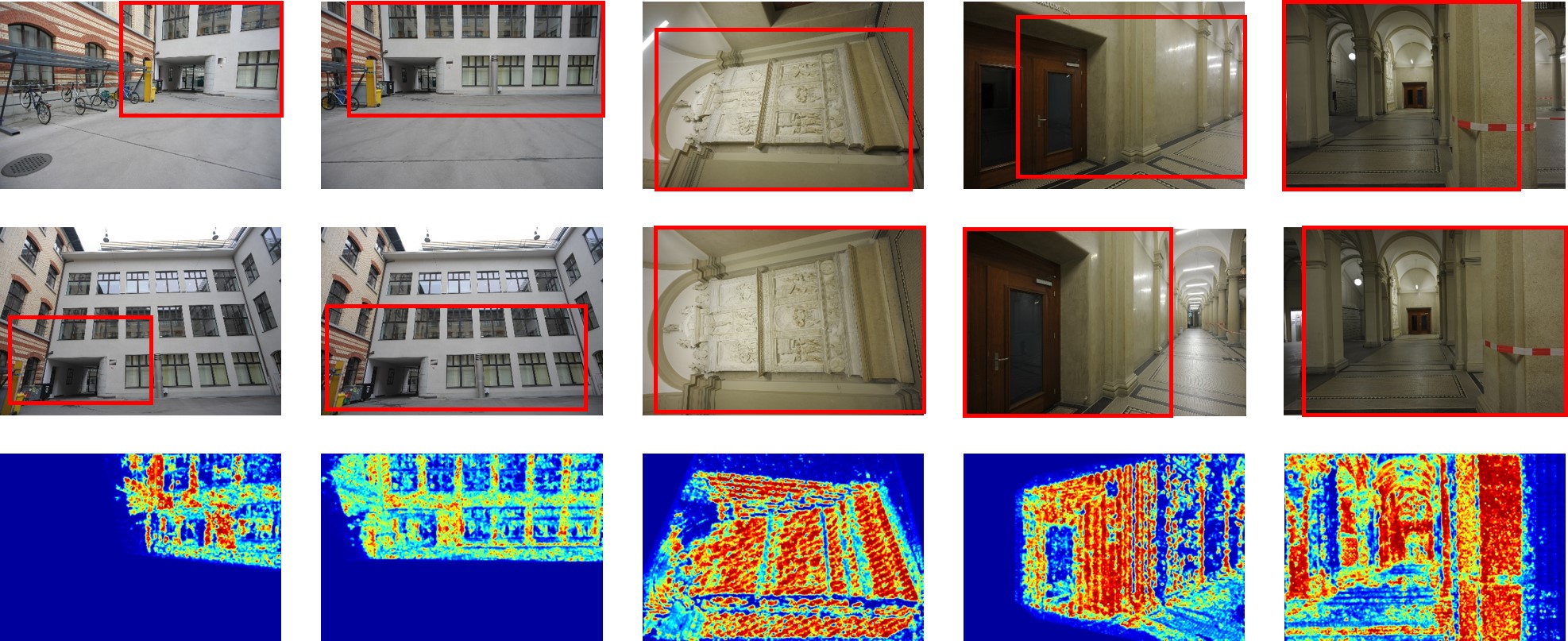}}
\caption{Visualization of our learned pixel-wise view weight on ETH3D~\cite{2017eth3d}. Top row: reference images. Middle row: source images. Bottom row: pixel-wise view weight. Areas marked with boxes in reference images and source images are co-visible. }
\label{fig:visual_view_weight}
\end{figure*}

Several examples for our estimated pixel-wise view weights are depicted in Fig.~\ref{fig:visual_view_weight}. 
Comparing the view weights with the visible areas in the reference validates that the 
all visible areas receive higher weights, while occluded and invisible parts have very low weights. 
Interestingly, in the first two images, pixels on the windows have low weights. 
Here, especially the upper row of windows mirror the surrounding buildings and 
cannot provide reasonable matching information. 
The other images have low weights in visible regions at areas with strong perspective and specular reflections as well as occlusions, while fronto-parallel and textured regions achieve higher weights.
We conclude that our pixel-wise view weight is capable to determine co-visible areas between the reference and source images.

\section{Visualization of Point Clouds}
\begin{figure*}[ht]
\centering
{\includegraphics[width=1.0\linewidth]{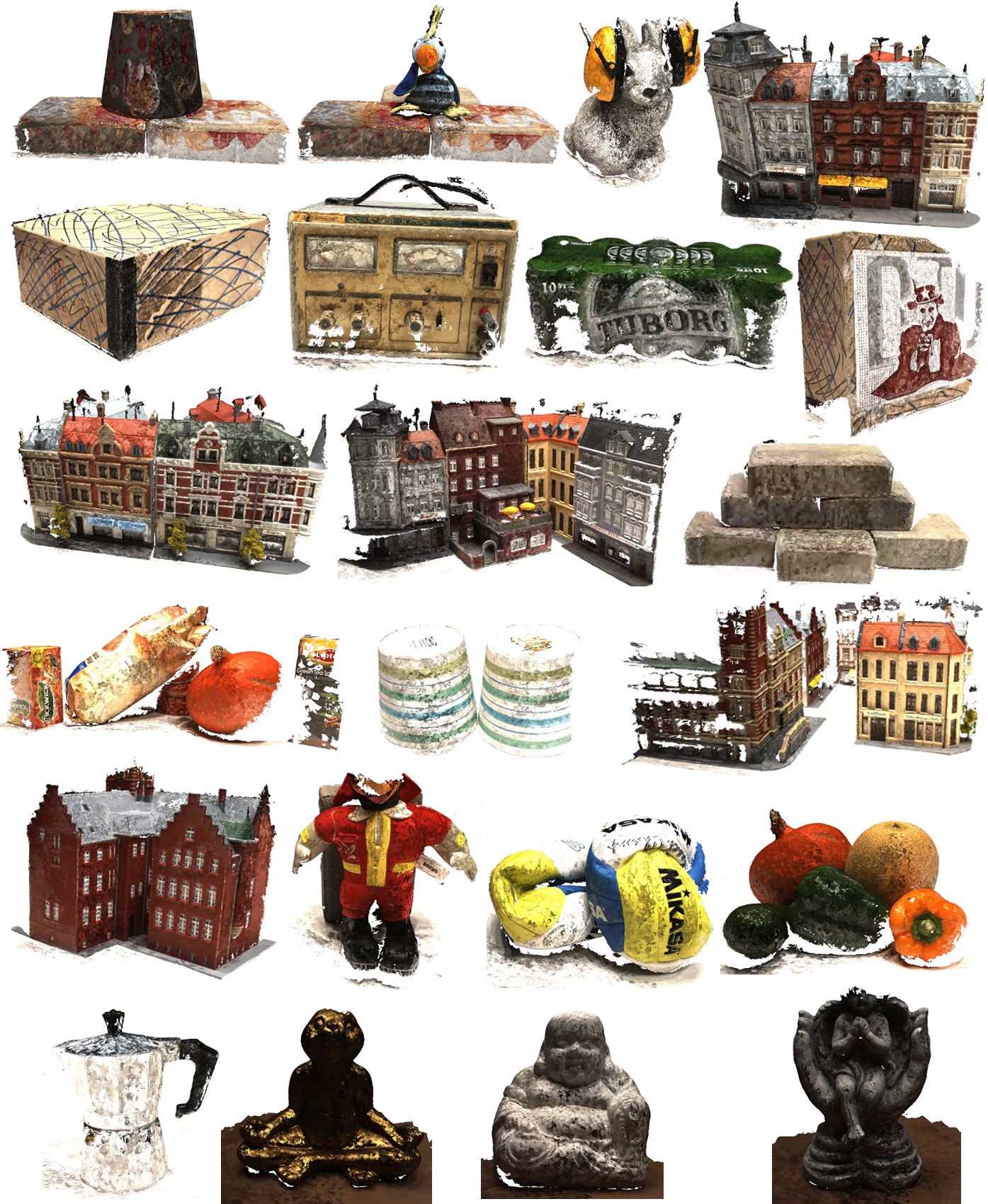}}
\caption{Reconstruction results on DTU's evaluation set~~\cite{aanaes2016_dtu}.}
\label{fig:visual_dtu}
\end{figure*}

\begin{figure*}
\centering
\begin{subfigure}[b]{\textwidth}
\centering
\includegraphics[width=0.9\textwidth]{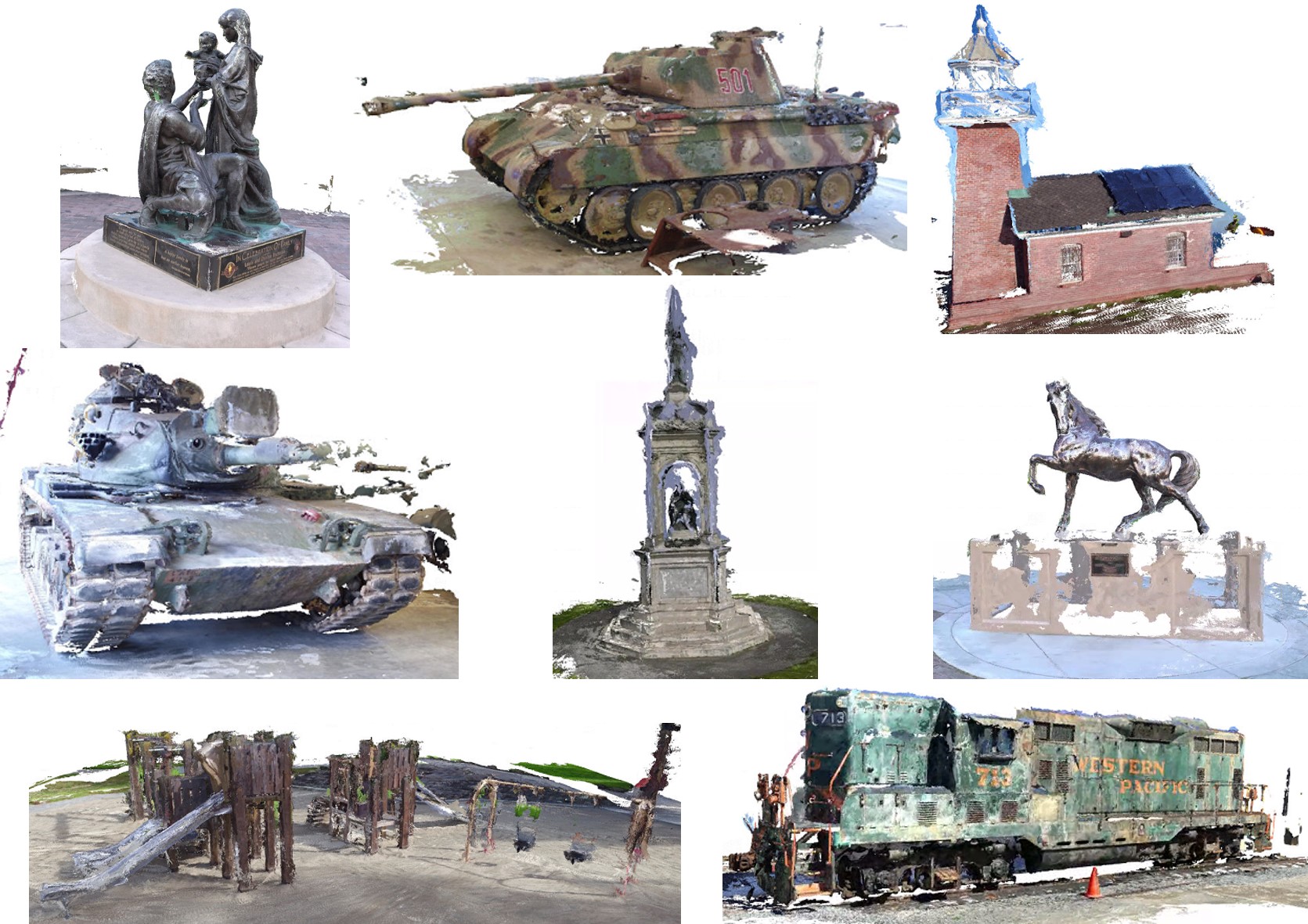}
\caption{Intermediate dataset}
\label{fig:tank_intermediate}
\end{subfigure}
\quad

\begin{subfigure}[b]{\textwidth}
\centering
\includegraphics[width=0.9\textwidth]{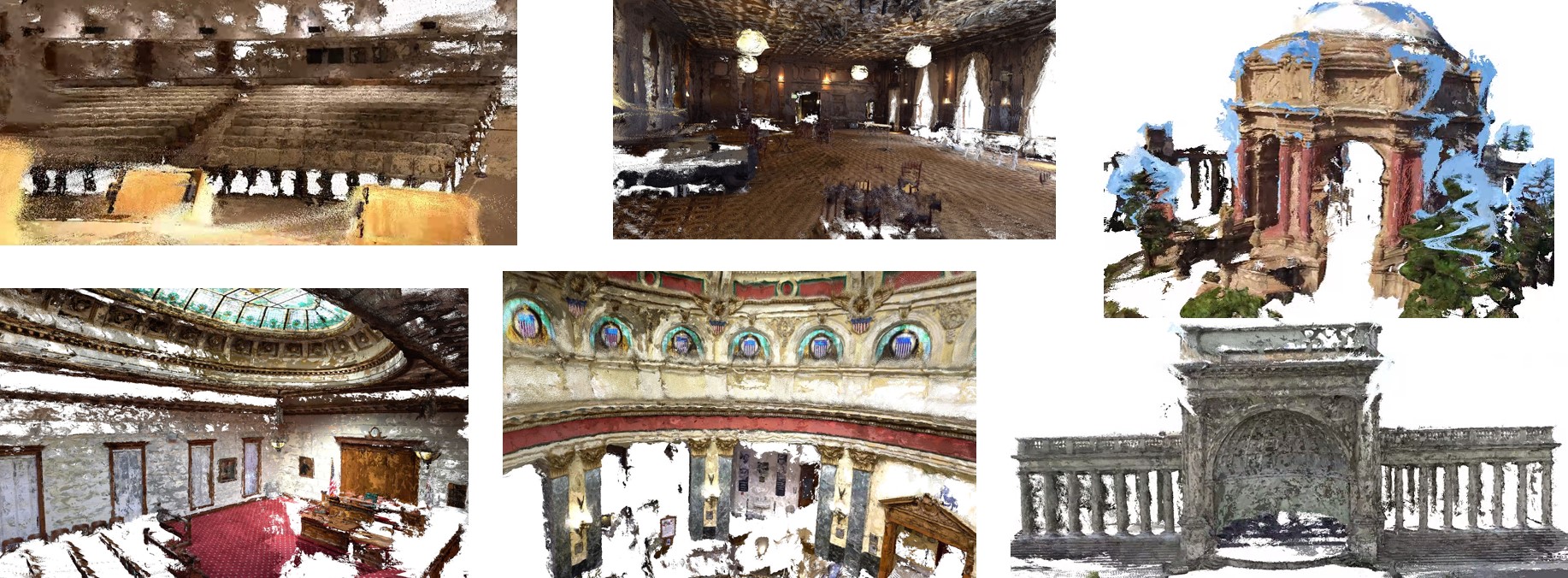}
\caption{Advanced dataset}
\label{fig:tank_advanced}
\end{subfigure}

\caption{Reconstruction results on Tanks \& Temples dataset~\cite{knapitsch2017tanks}.}
\label{fig:visual_tanks}
\end{figure*}

\begin{figure*}
\centering
\begin{subfigure}[b]{\textwidth}
\centering
\includegraphics[width=0.9\textwidth]{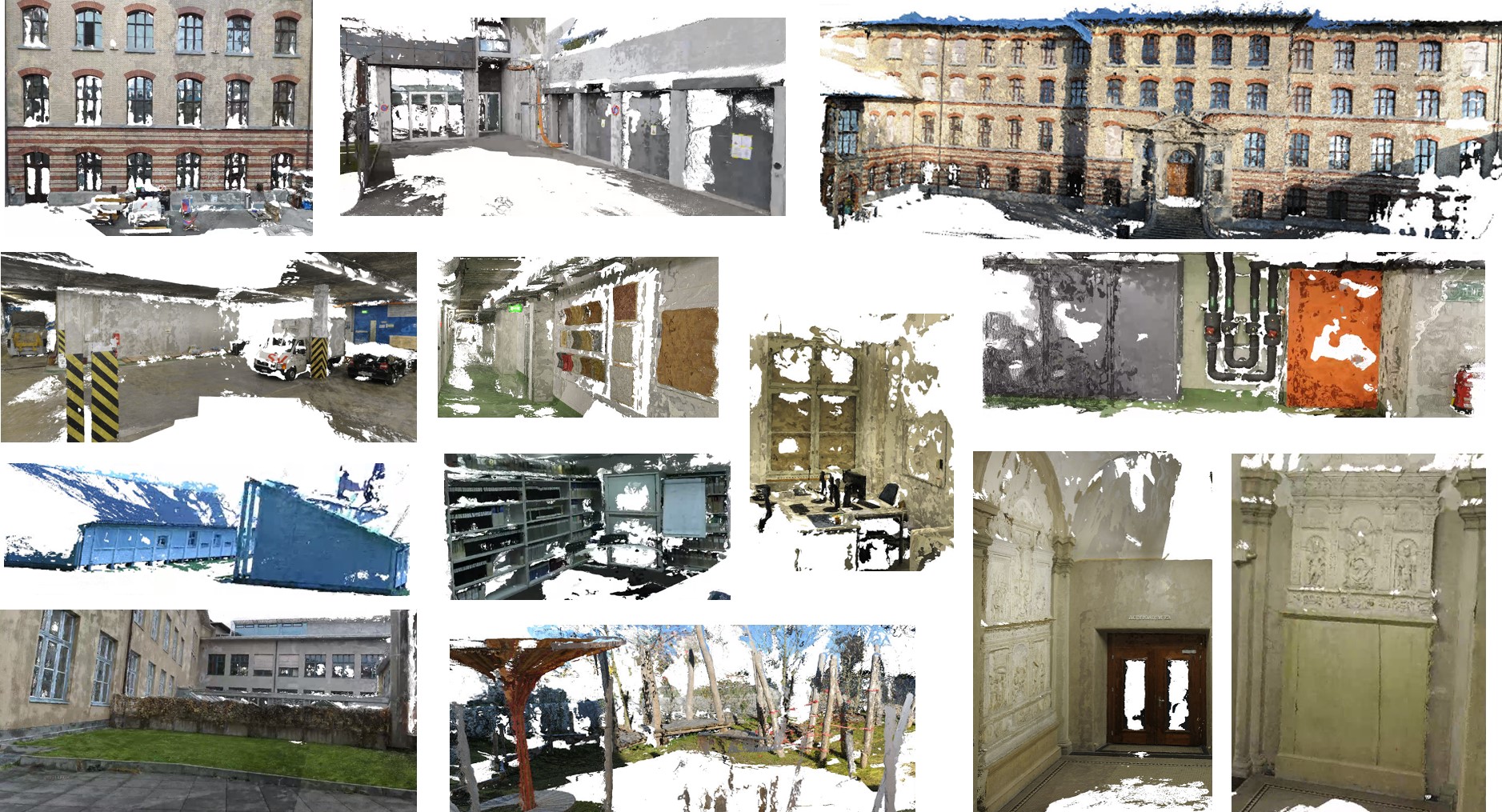}
\caption{Training dataset}
\label{fig:eth_training}
\end{subfigure}
\quad

\begin{subfigure}[b]{\textwidth}
\centering
\includegraphics[width=0.9\textwidth]{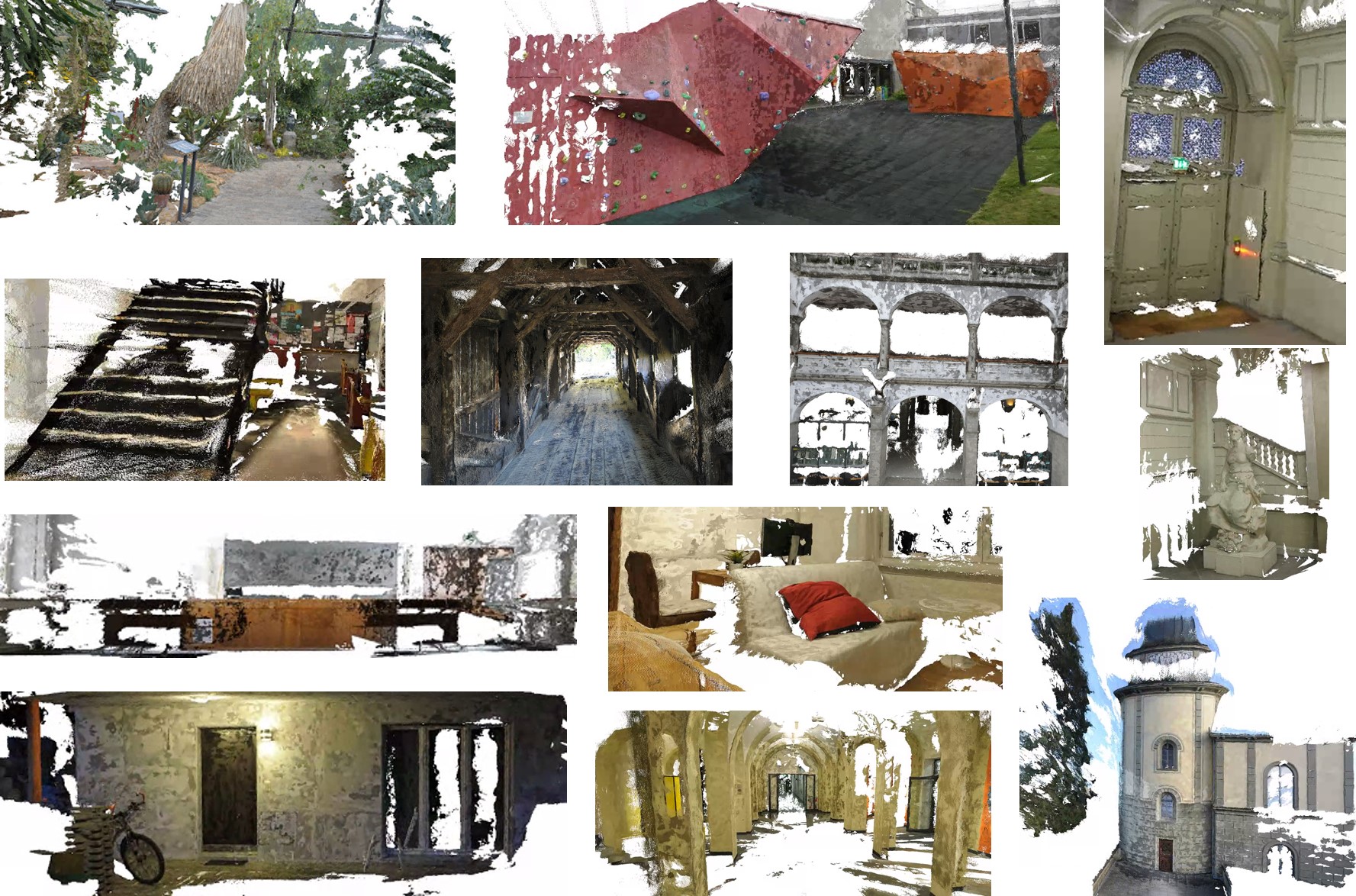}
\caption{Test dataset}
\label{fig:eth_test}
\end{subfigure}

\caption{Reconstruction results on ETH3D Benchmark~\cite{2017eth3d}.}
\label{fig:visual_eth}
\end{figure*}

We visualize the reconstructed point clouds from DTU's evaluation set~\cite{aanaes2016_dtu}, Tanks \& Temples dataset~\cite{knapitsch2017tanks} and ETH3D benchmark~\cite{2017eth3d} in Fig.~\ref{fig:visual_dtu}, \ref{fig:visual_tanks} and \ref{fig:visual_eth}.

\section{Future Work}
Currently, the learned confidence is only used to filter out unreliable estimates before depth fusion. 
However, we believe it will be a promising direction to further exploit the confidence in each GRU iteration. 
For example, one can refine the depth of unconfident areas with the information propagated from those confident areas~\cite{li2020confidence, kuhn2020deepc}. 
Another idea would be to focus more effort on unconfident areas only, while leaving the confident areas unchanged, which further should improve efficiency.

